\pdfoutput=1

\documentclass[11pt]{article}

\usepackage[]{ACL2024}

\usepackage{times}
\usepackage{latexsym}

\usepackage[T1]{fontenc}

\usepackage[utf8]{inputenc}

\usepackage{microtype}

\usepackage{inconsolata}

\usepackage{float}
\usepackage{multirow}
\usepackage{multicol}
\usepackage{amssymb}
\usepackage{pifont}

\usepackage{tcolorbox}
\usepackage{xcolor}
\usepackage{colortbl}
\usepackage{subfloat}
\usepackage{booktabs}
\usepackage{amsmath}
\usepackage{subfloat}
\usepackage{xspace}
\usepackage{subcaption}
\usepackage{graphics}
\usepackage{soul}

\usepackage{enumitem}
\setitemize{noitemsep,topsep=0pt,parsep=0pt,partopsep=0pt}

\newcommand{\blue}[1]{\textcolor{blue}{#1}}

\newcommand{\model}{LLoVi}

\newcommand{\cmark}{\ding{51}}%
\newcommand{\xmark}{\ding{55}}%

\title{A Simple LLM Framework for Long-Range Video Question-Answering}

\author{
Ce Zhang$^{*}$\ \ \ \ \ 
Taixi Lu$^{*}$\ \ \ \ \ 
Md Mohaiminul Islam\ \ \ \ \ 
Ziyang Wang\ \ \ \ \ \\
\textbf{Shoubin Yu}\ \ \ \ \ 
\textbf{Mohit Bansal}\ \ \ \ \ 
\textbf{Gedas Bertasius} \\
Department of Computer Science, UNC Chapel Hill \\
\tt cezhang@cs.unc.edu, taixi@email.unc.edu \\
\tt \{mmiemon, ziyangw, shoubin, mbansal, gedas\}@cs.unc.edu \\
\url{https://sites.google.com/cs.unc.edu/llovi}
}

\begin{document}
\maketitle
\begin{abstract}
We present \model, a simple yet effective \textbf{L}anguage-based \textbf{Lo}ng-range \textbf{Vi}deo question-answering (LVQA) framework. 
Our method decomposes the short- and long-range modeling aspects of LVQA into two stages. First, we use a short-term visual captioner to generate textual descriptions of short video clips (0.5-8 seconds in length) densely sampled from a long input video. Afterward, an LLM aggregates the densely extracted short-term captions to answer a given question.  
Furthermore, we propose a novel multi-round summarization prompt that asks the LLM first to summarize the noisy short-term visual captions and then answer a given input question. To analyze what makes our simple framework so effective, we thoroughly evaluate various components of our framework. Our empirical analysis reveals that the choice of the visual captioner and LLM is critical for good LVQA performance. The proposed multi-round summarization prompt also leads to a significant LVQA performance boost.
Our method achieves the best-reported results on the EgoSchema dataset, best known for very long-form video question-answering. \model~also outperforms the previous state-of-the-art by \textbf{10.2\%} and \textbf{6.2\%} on NExT-QA and IntentQA for LVQA.
Finally, we extend \model~to grounded VideoQA, which requires both QA and temporal localization,
and show that it outperforms all prior methods on NExT-GQA. Code is available at \url{https://github.com/CeeZh/LLoVi}.
\end{abstract}
    
\section{Introduction}

\begin{figure}[t]
\hspace{2cm}
    \centering
    \hfill
    \includegraphics[width=1.0\linewidth]{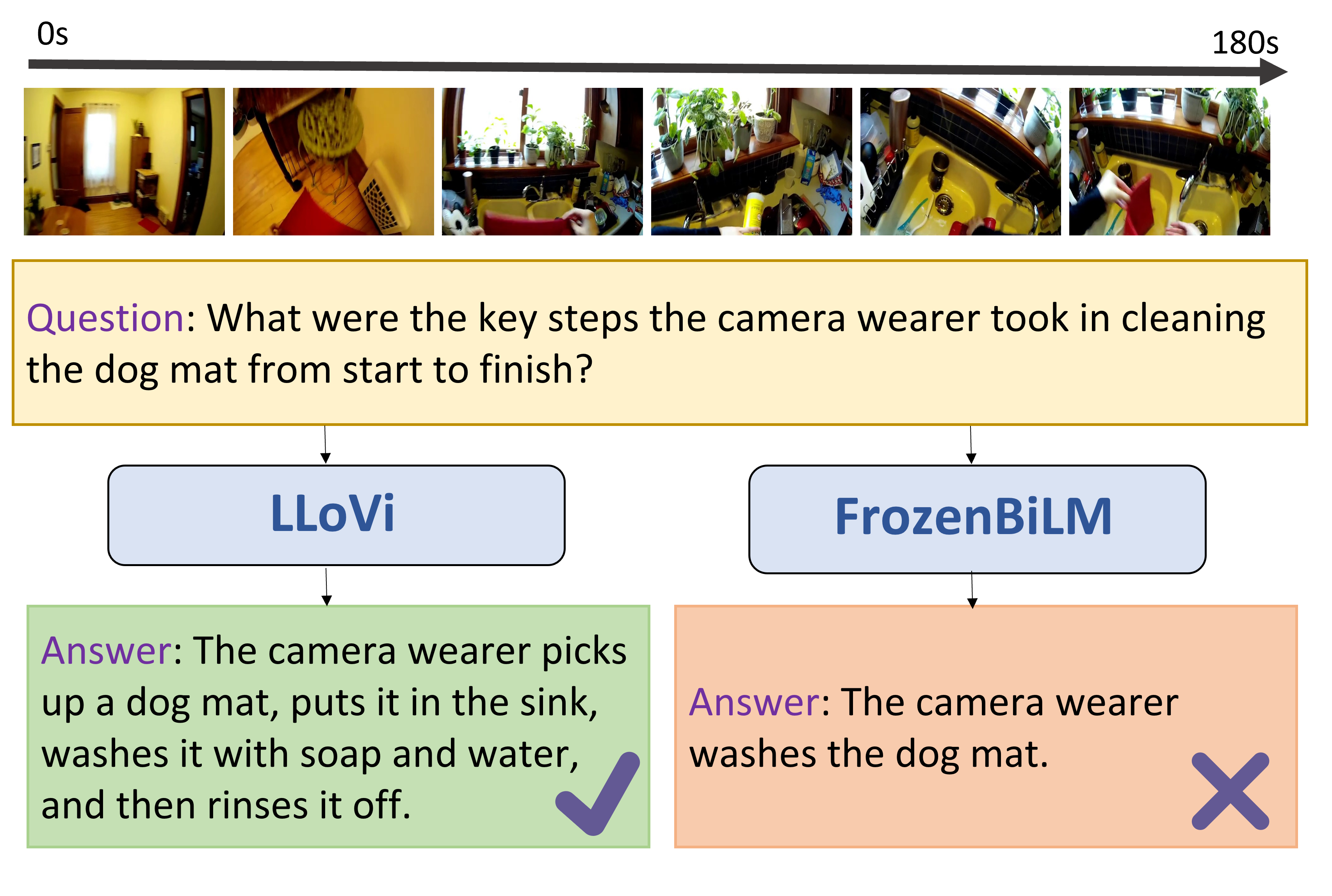}
    \vspace{-8mm}
    \caption{Comparison between LLoVi (ours) and the recent FrozenBiLM~\cite{yang2022frozenbilm} video QA method. Like most prior methods, FrozenBiLM is best suited for short-range video understanding. Thus, as illustrated in the figure, it fails to answer a question that requires reasoning about complex human activities in a long video. In comparison, our method effectively reasons over long temporal extents and produces a correct answer.}
    \vspace{-0.1in}
    \label{fig:acc}
\end{figure}

Recent years have witnessed remarkable progress in short video understanding (5-15s in length)~\cite{wang2022internvideo,ye2023mplug,fu2021violet,yang2022frozenbilm, wang2023unified}. However, extending these models to long videos (e.g., several minutes or hours in length) is not trivial due to the need for sophisticated long-range temporal reasoning capabilities. Most existing long-range video models rely on costly and complex long-range temporal modeling schemes, which include memory queues~\cite{wu2022memvit,chen2020memory,lee2021video, lee2018memory},
long-range feature banks~\cite{wu2019long,cheng2022tallformer,zhang2021temporal},
space-time graphs~\cite{hussein2019videograph,wang2021supervoxel},
state-space layers~\cite{islam2022long,islam2023efficient,wang2023selective}
and other complex long-range modeling modules~\cite{hussein2019timeception,bertasius2021space,yang2023relational}.

Recently, Large Language Models (LLMs) have shown impressive capability for long-range reasoning on a wide range of tasks such as document understanding~\cite{sun2023pearl, wang2023gpt, gur2023real} and long-horizon planning~\cite{liu2023llm,hao2023reasoning,song2023llm}. Motivated by these results in the natural language and decision-making domain, we explore using LLMs for long-range video question answering (LVQA). Specifically, we propose \model, a simple yet effective language-based framework for long-range video understanding. Unlike prior long-range video models, our approach does not require specialized long-range video modules (e.g., memory queues, state-space layers, etc.) but instead uses a short-term visual captioner coupled with an LLM, thus exploiting the long-range temporal reasoning ability of LLMs. Our simple two-stage framework tackles the LVQA task by decomposing it into short and long-range modeling subproblems: 
\begin{enumerate}[noitemsep,topsep=0pt,parsep=0pt,partopsep=0pt,leftmargin=15pt]
    \item First, given a long video input, we segment it into multiple short clips and convert them into short textual descriptions using a pre-trained frame/clip-level visual captioner (e.g., BLIP2~\cite{li2023blip2}, LaViLa~\cite{zhao2023learning}, LLaVA~\cite{liu2023llava}).
    \item Afterwards, we concatenate the temporally ordered captions from Step 1 and feed them into an LLM (e.g., GPT-3.5, GPT-4, LLaMA) to perform long-range reasoning for LVQA.
\end{enumerate}

To further enhance the effectiveness of our framework, we also introduce a novel multi-round summarization prompt that asks the LLM first to summarize the short-term visual captions and then answer a given question based on the LLM-generated video summary. Since the generated captions may be noisy or redundant, such a summarization scheme enables filtering out potentially distracting/irrelevant information and eliminating redundant sentences, which significantly improves the reasoning ability of the LLM for LVQA.

Additionally, we conduct an empirical study on EgoSchema to investigate the factors behind our framework's success. Specifically, we study (i) the selection of a visual captioner, (ii) the choice of an LLM,  (iii) the LLM prompt design, and (iv) optimal video processing configurations. 
Our key empirical findings include:
{\setlength{\parindent}{0em}
\begin{itemize}[leftmargin=10pt]
   \item Our newly proposed multi-round summarization prompt leads to the most significant boost in performance (\textbf{+3.6\%}) among the prompts we have tried (e.g., Chain-of-Thought, Plan-and-Solve).
  \item GPT-4 as an LLM provides the best performance, while GPT-3.5 provides the best trade-off between the accuracy and the cost.
  \item LaViLa~\cite{zhao2023learning} as a visual captioner produces best results (\textbf{55.2\%}) followed by BLIP-2~\cite{li2023blip2} (\textbf{50.6\%}) and EgoVLP~\cite{qinghong2022egocentric} (\textbf{46.6\%}).
  \item Extracting visual captions from consecutive 1-second video clips of the long video input leads to the best results. Also, extracting captions from sparsely sampled video clips leads to \textbf{8x} improved efficiency while still maintaining reasonable performance (\textbf{2.0\%} accuracy drop).
\end{itemize}
}

\noindent We want to make it clear that \model~is not based on any complex or novel design choices. It is a simple, effective, and training-free method that outperforms all prior approaches on EgoSchema, NExT-QA, IntentQA, and NeXT-GQA, establishing a strong baseline for the LVQA task. We hope that our work will encourage the LVQA community to build on our work and use our thorough empirical insights to develop new LVQA models. 

\section{Related Work}
\noindent \textbf{Long-range Video Understanding.} Modeling long-range videos (e.g., several minutes or longer) typically requires models with sophisticated temporal modeling capabilities, often leading to complex model design. LF-VILA~\cite{sun2022long} proposes a Temporal Window Attention (HTWA) mechanism to capture long-range dependency in long-form video. MeMViT~\cite{wu2022memvit} and MovieChat~\cite{song2023moviechat} adopt a memory-based design to store information from previously processed video segments. Several prior methods use space-time graphs~\cite{hussein2019videograph,wang2021supervoxel} or relational space-time modules~\cite{yang2023relational} to capture spatiotemporal dependencies in long videos. Lastly, the recently introduced S4ND~\cite{nguyen2022s4nd}, ViS4mer~\cite{islam2022long} and S5~\cite{wang2023selective} use Structured State-Space Sequence (S4)~\cite{gu2021efficiently} layers to capture long-range dependencies in the video. Unlike these prior approaches, we do not use any complex long-range temporal modeling modules but instead develop a simple and strong LLM-based framework for zero-shot LVQA.

\noindent \textbf{LLMs for Video Understanding.} 
The recent surge in large language models (LLMs)~\cite{brown2020language, openai2023gpt4,touvron2023llama, raffel2020exploring, chung2022scaling, tay2022ul2} has inspired many LLM-based applications in video understanding.  Methods like Socratic Models~\cite{zeng2022socraticmodels} and VideoChat~\cite{li2023videochat} integrate pretrained visual models with LLMs for extracting visual concepts and applying them to video tasks. Video ChatCaptioner~\cite{chen2023video} and ChatVideo~\cite{wang2023chatvideo} leverage LLMs for video representation and dialog-based user interaction, respectively. VidIL~\cite{wang2022language} employs LLMs for adapting image-level models to video tasks using few-shot learning. Beyond short-term video understanding, the works in~\cite{2023mmvid,chung2023long, bhattacharya2023video} explored LLMs for long-range video modeling. The work in~\cite{2023mmvid} uses GPT-4 for various long-range video modeling tasks but lacks quantitative evaluation. Meanwhile,~\cite{chung2023long} focuses on movie datasets, requiring limited visual analysis~\cite{mangalam2023egoschema} and mostly relying on non-visual speech/subtitle inputs. In contrast to these prior methods, we focus on the LVQA task and provide an extensive empirical analysis of various design choices behind our LLM framework.

\noindent \textbf{Video Question Answering.} 
Unlike image question-answering, video question-answering (VidQA) presents unique challenges, requiring both spatial and temporal reasoning. Most existing VidQA methods, either using pretraining-finetuning paradigms~\cite{VindLU_CVPR2023,lei2021less,yu2023self}, zero-shot~\cite{yang2022zero,suris2023vipergpt,lin2023mm,yu2023self}, or few-shot learning~\cite{wang2022language}, focus on short-term video analysis (5-30s). To overcome the limitations of short-term VidQA, new benchmarks have been proposed: ActivityNet-QA~\cite{yu2019activitynet}, TVQA~\cite{lei2018tvqa}, How2QA~\cite{yang2021just}, MovieQA~\cite{tapaswi2016movieqa}, and DramaQA~\cite{DBLP:conf/aaai/ChoiOHSJLZ21} ranging from 100s to several minutes in video duration. Despite longer video lengths, the analysis in ~\cite{mangalam2023egoschema, yang2020gives,jasani2019we} found that many of these benchmarks can be solved by analyzing only short clips (i.e., not requiring long-range video modeling) or by using pure text-only methods that ignore visual content. To address these issues, the EgoSchema benchmark~\cite{mangalam2023egoschema} was recently introduced, requiring at least 100 seconds of video analysis and not exhibiting language-based biases. 

\begin{figure}[t]
    \centering
    \includegraphics[width=1.0\linewidth]{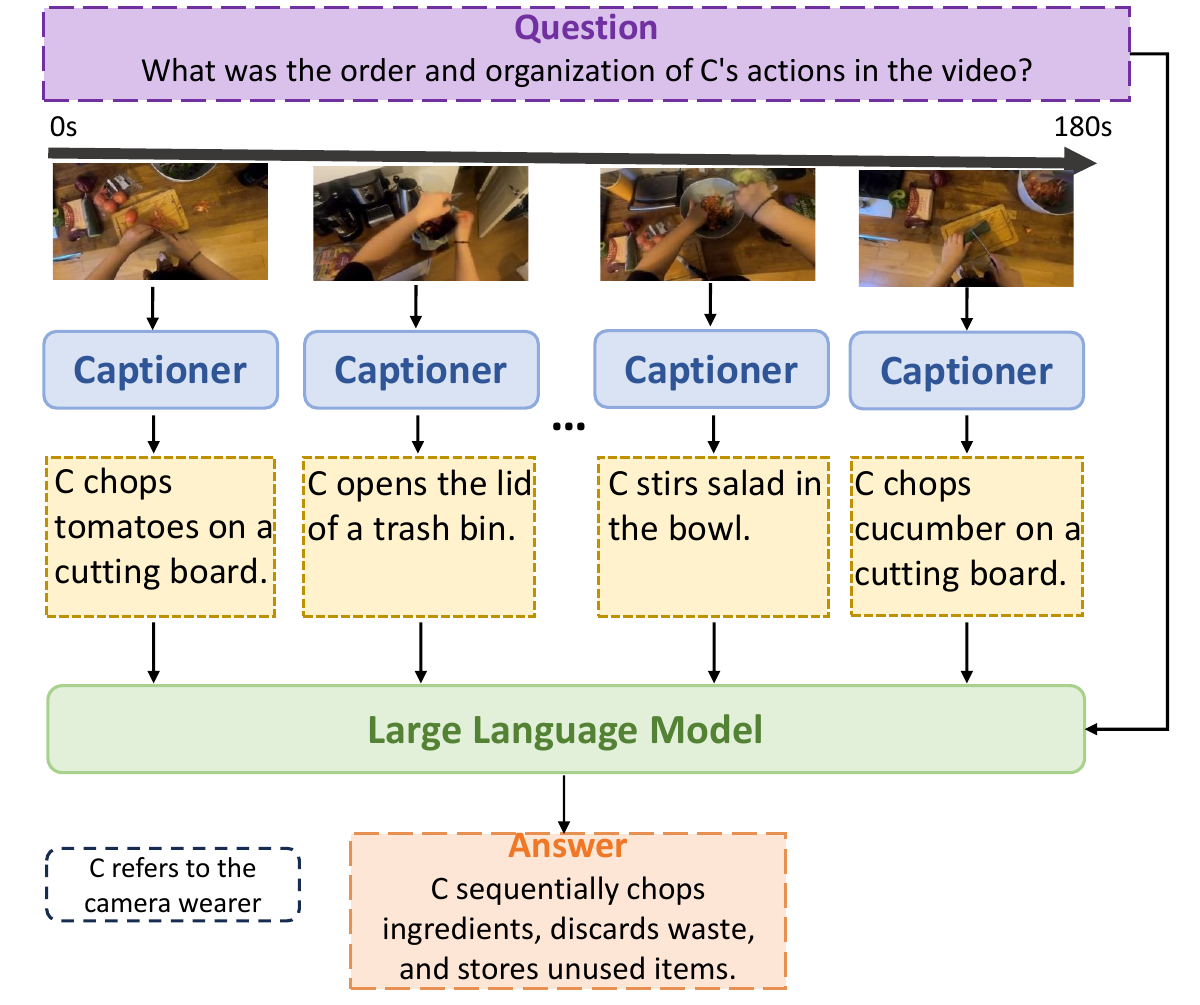}
    \vspace{-0.2in}
    \caption{An illustration of LLoVi, our simple LLM framework for long-range video question-answering (LVQA). We use Large Language Models (LLMs) like GPT-3.5 and GPT-4 for their long-range modeling capabilities. Our method involves two stages: first, we use short-term visual captioners (e.g, LaViLa, BLIP2) to generate textual descriptions for brief video clips (0.5s-8s). Then, an LLM aggregates these dense, short-term captions for long-range reasoning required for LVQA. This simple approach yields impressive results, demonstrating LLMs' effectiveness in LVQA.}
    \label{fig:teaser}
\end{figure}

\noindent \textbf{LLM Prompt Design.} With the emergence of LLMs, there has been an increasing research emphasis on LLM prompt design. The recent works in \cite{wei2022chain, zhou2022least, schick2020exploiting, chen2022program, yao2022react} explored prompting strategy in few-shot learning settings. To eliminate the need for extensive human annotations, \cite{kojima2022large, wang2023plan, wang2022self} proposed zero-shot prompting methods. Subsequent research~\cite{zhou2022large, zhang2022automatic, pryzant2023automatic} has concentrated on the automatic refinement of prompts. Instead, we propose a multi-round summarization LLM prompt for handling long, noisy, and redundant textual inputs describing video content for LVQA.

\section{Method}
\label{sec:method}

\begin{figure*}[htbp]
    \centering
    \includegraphics[width=1.0\linewidth]{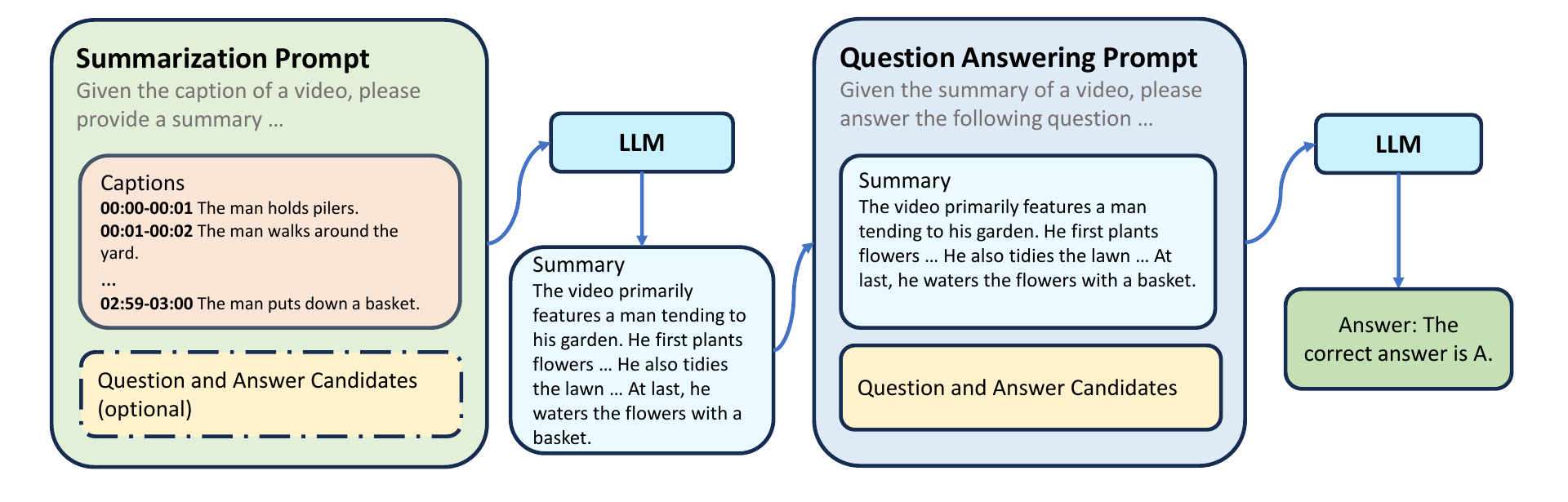}
    \vspace{-0.3in}
    \caption{
    An illustration of our multi-round summarization prompt that first asks an LLM  to summarize the noisy short-term visual captions (first round of prompting) and then answer a given question about the video based on the LLM-generated summary (second round of prompting). Our results indicate that such a multi-round prompting strategy significantly boosts LVQA performance compared to standard prompting techniques (\textbf{+5.8\%}).}
    \label{fig:Prompt Design}
\end{figure*}

Our method, LLoVi, decomposes LVQA into two subtasks: 1) short-term video clip captioning and 2) long-range text-based video understanding. Our decomposed LVQA framework brings several important advantages. First, our approach is simple as it does not rely on complex/specialized long-range video modeling operators (e.g., memory queues, state-space layers, space-time graphs, etc.). Second, our framework is training-free, which makes it easy to apply it to LVQA in zero-shot settings. Third, our framework enables us to leverage the strong existing short-term visual captioners (e.g., LaViLa, LLaVA) and powerful zero-shot LLMs (e.g., GPT-3.5, GPT-4, LLaMA). Fourth, our method is highly flexible, i.e., it can incorporate various visual captioners and LLMs, and also benefit from future improvements in visual captioning/LLM model design. Figure~\ref{fig:teaser} presents a detailed illustration of our high-level approach. Below, we provide details about each component of our framework.

\subsection{Short-term Video Clip Captioning}

Given a long untrimmed video input $V$, we first segment it into $N_v$ non-overlapping short video clips $v = \{v_m\}_{m=1}^{N_v}$, where $v_m \in \mathbb{R}^{T_v \times H \times W \times 3}$ and $T_v, H, W$ are the number of frames, height and width of a short video clip respectively. Afterward, we feed each video clip $v_m$ into a pretrained short-term visual captioner $\phi$, which produces textual captions $c_m = \phi(v_m)$, where $c_m = (w_1, \hdots, w_{L_m})$ and $w_i$ represents the i-th word in caption $c_m$ of length $L_m$. Note that our model is not restricted to any specific visual captioning model. Our experimental section demonstrates that we can incorporate various video (LaViLa~\cite{zhao2023learning}, EgoVLP~\cite{qinghong2022egocentric}, and image (BLIP-2~\cite{li2023blip}) captioning models. Next, we describe how our extracted short-term captions are processed by an LLM.

\subsection{Long-range Reasoning with an LLM}

We want to leverage foundational LLMs for holistic long-range video understanding. Formally, given short-term visual captions $\{c_m\}_{m=1}^{N_v}$ for all $N_v$ short video clips, we first concatenate the clip captions into the full video captions $C = [ c_1, \hdots, c_{N_v} ]$ in the same order as the captions appear in the original video. Afterward, the concatenated video captions $C$ are fed into an LLM for long-range video reasoning. Specifically, given the concatenated video captions $C$, the question $Q$, and the answer candidates $A$, we prompt the LLM to select the correct answer using the following prompt template: \textit{``Please provide a single-letter answer (A, B, C, D, E) to the following multiple-choice question $\{Q\}$. You are given language descriptions of a video. Here are the descriptions: $\{C\}$. Here are the choices $\{A\}$."}. The full prompt is included in the Supplementary Material. 

Our experiments in Section~\ref{main_result} suggest that this simple approach works surprisingly well for LVQA. However, we also discovered that many modern LLMs (e.g., GPT-3.5, LLaMA) may struggle when provided with long ($>$1K words), noisy, and potentially redundant/irrelevant caption sequences. To address these issues, we investigate more specialized LLM prompts that ask an LLM first to summarize the noisy short-term visual captions (first round of prompting) and then answer a given question about the video (second round of prompting). Specifically, we formulate such a multi-round prompt as follows:  given the video captions $C$, the question $Q$, and the answer candidates $A$, instead of directly feeding the $\{C, Q, A\}$ triplet into LLM for LVQA, we first ask the LLM to provide a summary of the captions in the first round, which we denote as $S$ using the following prompt template: ``\textit{You are given language descriptions of a video: \{$C$\}. Please give me a \{$N_{w}$\} word summary.}" $N_w$ denotes the desired number of words in the summary $S$. 
Afterward, during the second round of prompting, instead of using the captions $C$, we use the summary $S$ as input for the LLM to select one of the answer candidates. Conceptually, such a prompting scheme is beneficial, as the LLM-generated summary $S$ filters out irrelevant/noisy information from the initial set of captions $C$, making LLM inputs for the subsequent QA process more succinct and cleaner. A detailed illustration of our multi-round prompt is shown in Figure~\ref{fig:Prompt Design}.

\subsection{Implementation Details}
For the experiments on EgoSchema, we use LaViLa~\cite{zhao2023learning} as our captioner. We segment each video into multiple 1s clips, resulting in a list of consecutive clips that cover the entire video. We use GPT-3.5 as the LLM on EgoSchema. For NExT-QA, IntentQA, and NExT-GQA, we use CogAgent~\cite{hong2024cogagent} as the visual captioner and GPT-4 as the LLM. We downsample the videos to 0.5 FPS and prompt CogAgent to generate captions for each frame.
More details are provided in the Supplementary Material.

\section{Experiments}

\begin{figure}[t]
    \centering
    \hfill
    \includegraphics[width=1.0\linewidth]{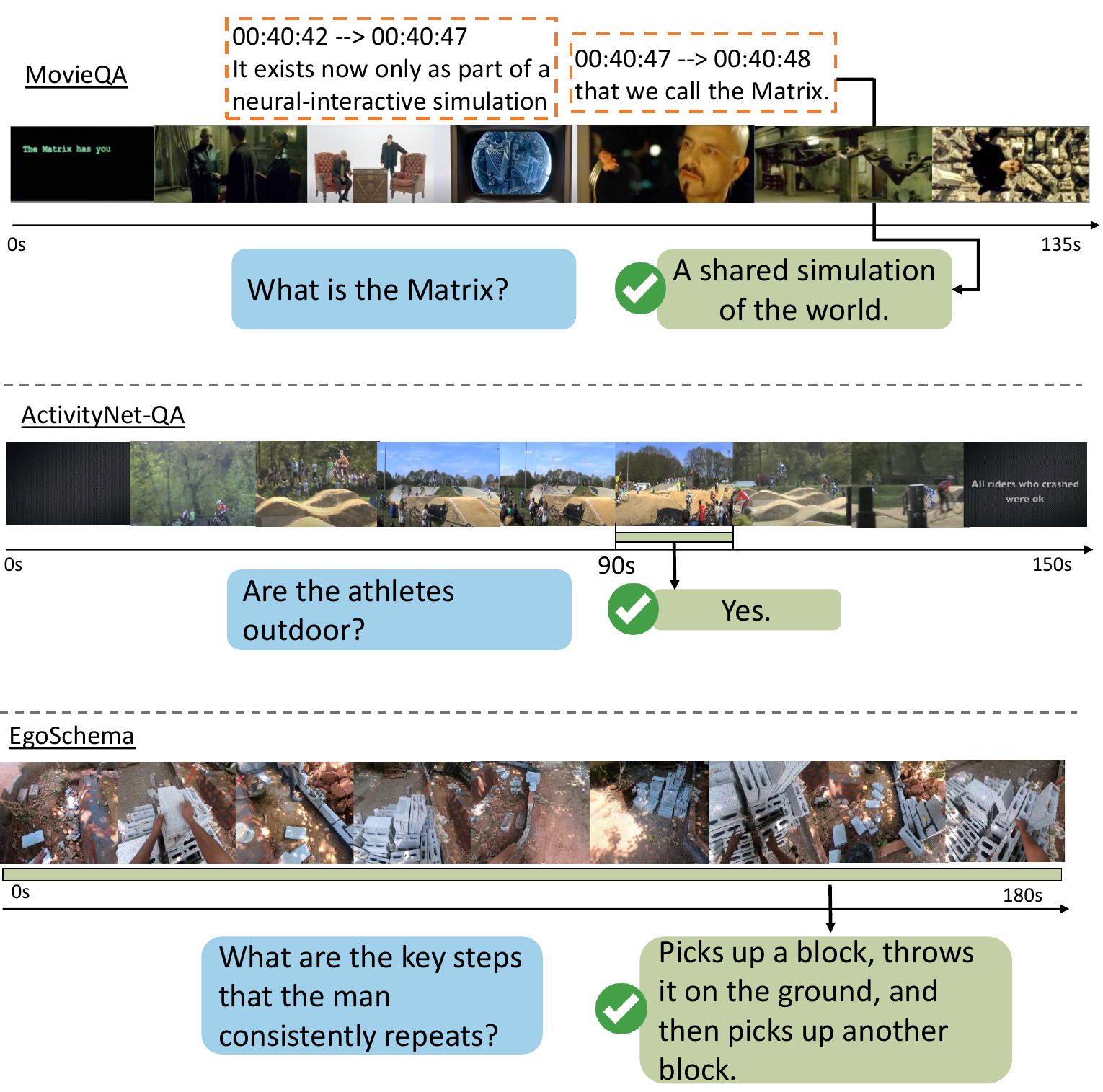}
    \vspace{-0.2in}
    \caption{\textbf{An illustration of prior LVQA dataset limitations.} \textbf{Top}: An example from MovieQA~\cite{tapaswi2016movieqa}. The model can use the provided subtitle information to answer a question while ignoring visual cues in a video. \textbf{Middle}: An example from the ActivityNet-QA Dataset~\cite{yu2019activitynet}. Despite long video inputs, the model only needs to analyze a short 1s video clip to answer the question. \textbf{Bottom}: An example from the EgoSchema Dataset~\cite{mangalam2023egoschema}. The model must analyze visual cues from the video to answer a given question without relying on additional textual inputs (e.g., speech, subtitles).}
    \label{fig:dataset}
\end{figure}
    
\subsection{Datasets and Metrics}
Unlike short-term video question-answering, long-range video question-answering (LVQA) lacks robust and universally agreed-upon benchmarks.  
As shown in Figure~\ref{fig:dataset}, many prior LVQA benchmarks either exhibit significant language biases, or do not require long-range video modeling capabilities. 
To address these limitations, recent work introduced \textbf{EgoSchema}~\cite{mangalam2023egoschema}, a new long-range video question-answering benchmark consisting of 5K multiple choice question-answer pairs spanning 250 hours of video and covering a wide range of human activities. By default, our experiments are conducted on the validation set of 500 questions (referred to as the EgoSchema Subset). The final comparison is done on the full test set of 5K EgoSchema questions. We use QA accuracy (i.e., the percentage of correctly answered questions) as our evaluation metric. Additionally, we also perform zero-shot LVQA experiments on three commonly-used LVQA benchmarks: \textbf{NExT-QA}~\cite{xiao2021next}, \textbf{IntentQA}~\cite{li2023intentqa}, and \textbf{NExT-GQA}~\cite{xiao2023can}. Detailed dataset information and metrics can be found in the Supplementary Material.

\subsection{Empirical Study on EgoSchema}
We first study the effectiveness of different components within our \model~framework, including (i) the visual captioner, (ii) the LLM, (iii) the optimal video processing configurations, and (iv) the LLM prompt design. The experiments are conducted on the EgoSchema Subset. We discuss our empirical findings below. We also include additional experiments in the Supplementary Material.

\subsubsection{Visual Captioning Model}

\begin{table}[t]
\centering
\setlength{\tabcolsep}{1pt}
\resizebox{\columnwidth}{!}{
\begin{tabular}{lcccc}
\toprule
\multirow{2}{*}{Captioner} & Caption & Ego4D & \multirow{2}{*}{Acc. (\%)} \\
& Type & Pre-training & \\
\midrule
EgoVLP~\cite{qinghong2022egocentric} & clip-level & \cmark & 46.6 \\
LLaVA~\cite{liu2023llava} & frame-level & \xmark & 45.2 \\
BLIP-2~\cite{li2023blip} & frame-level & \xmark & 50.6 \\
LaViLa~\cite{zhao2023learning} & clip-level & \cmark & \textbf{55.2} \\
\color{gray}Oracle & \color{gray}clip-level & - & \color{gray}66.0 \\
\bottomrule
\end{tabular}
}
\caption{\textbf{Accuracy of our framework with different visual captioners.} LaViLa visual captioner achieves the best results, outperforming other clip-level (e.g., EgoVLIP, VideoBLIP) and image-level (e.g., BLIP-2) captioners. We also observe that the Oracle baseline using ground truth captions greatly outperforms all other variants, suggesting that our framework can benefit from the future development of visual captioners.}
\label{tab:captioner}
\vspace{-0.1in}
\end{table}

In Table~\ref{tab:captioner}, we study the effectiveness of various clip-level video captioners, including LaViLa~\cite{zhao2023learning} and EgoVLP~\cite{qinghong2022egocentric}. In addition to video captioners, we also try the state-of-the-art image captioners, BLIP-2~\cite{li2023blip2} and LLaVA-1.5~\cite{liu2023llava}. Lastly, to study the upper bound of our visual captioning results, we include the ground truth Oracle captioning baseline obtained from the Ego4D dataset. All baselines in Table~\ref{tab:captioner} use similar experimental settings, including the same LLM model, i.e., GPT-3.5. The results are reported as LVQA accuracy on the EgoSchema Subset. 
Table~\ref{tab:captioner} suggests that LaViLa provides the best results, outperforming BLIP-2, EgoVLP, and LLaVA. We also observe that despite not being pre-trained on Ego4D~\cite{grauman2022ego4d}, BLIP-2 performs reasonably well (\textbf{50.6\%}) and even outperforms a strong Ego4D-pretrained baseline, EgoVLP. Lastly, the Oracle baseline with ground truth captions outperforms LaViLa captions by a large margin (\textbf{10.8\%}). This shows that our method can benefit from future improvements in captioning models.

In addition to our quantitative analysis, we also observed that our framework with the LaViLa captioner demonstrates basic Person Re-Identification capabilities when the video involves simple interactions among people. We visualize these results in our Supplementary Material.

\begin{table}[t]
\centering
\resizebox{\columnwidth}{!}{
    \begin{tabular}{lccccc}
    \toprule
    LLM & Model Size & Acc. (\%)\\ 
    \midrule
    Mistral~\cite{jiang2023mistral} & 7B & 50.8 \\
    Llama3-8B~\cite{touvron2023llama} & 8B & 52.2 \\
    Llama3-70B~\cite{touvron2023llama} & 70B & 56.8 \\
    GPT-3.5~\cite{gpt35} & 175B & 55.2 \\
    GPT-4~\cite{openai2023gpt4} & 1.8T & \textbf{61.2} \\
    \bottomrule
    \end{tabular}
}
\caption{\textbf{Accuracy of our framework with different LLMs.} GPT-4 achieves the best accuracy, suggesting that stronger LLMs perform better in LVQA. However, we use GPT-3.5 for most of our experiments due to the best accuracy and cost tradeoff.}
\vspace{-0.1in}
\label{tab:llm}
\end{table}

\subsubsection{Large Language Model}
In Table~\ref{tab:llm}, we analyze the performance of our framework using different LLMs while fixing the visual captioner to be LaViLa. Our results indicate that GPT-4 achieves the best performance (\textbf{61.2\%}), followed by LLama3-70B (\textbf{56.8\%}) and GPT-3.5 (\textbf{55.2\%}). Thus, stronger LLMs (GPT-4) are better at long-range modeling, as indicated by a significant margin in LVQA accuracy between GPT-4 and all other LLMs ($>$\textbf{4.4\%}).
We also observe that despite having a much smaller number of parameters, LLama3-8B (\textbf{52.2}\%) and Mistral-7B (\textbf{50.8}\%) still achieve competitive performance. Due to the high cost of GPT-4 and the large computational resource requirements of Llama3-70B, we use GPT-3.5 for most of our experiments unless noted otherwise.

\begin{table}[t]
\centering
\setlength{\tabcolsep}{12pt}
\resizebox{\columnwidth}{!}{
    \begin{tabular}{lcccc}
    \toprule
    Clip length (s) & 1 & 2 & 4 & 8 \\ 
    \midrule
    Acc. (\%) & \textbf{55.2} & 54.8 & 53.4 & 53.4\\
    \bottomrule
    \end{tabular}
    }
\caption{\textbf{Analysis of different clip length.} We divide the input long video into consecutive clips of different length. The highest accuracy is achieved when the clips are shortest, while performance diminishes as clip length increases. This indicates that splitting long videos into shorter segments, particularly 1-second clips, is the most efficient approach.}
\vspace{-0.1in}
\label{tab:clip_length}
\end{table}

\begin{table}[t]
\centering
\setlength{\tabcolsep}{12pt}
\resizebox{\columnwidth}{!}{
    \begin{tabular}{lcccc}
    \toprule
    Clip sampling rate & 1 & 1/2 & 1/4 & 1/8 \\ 
    \midrule
    Acc. (\%) & \textbf{55.2} & 55.2 & 54.6 & 53.2 \\
    \bottomrule
    \end{tabular}
    }
\caption{\textbf{Analysis of sparse video clip sampling. 
} We divide the input long video into consecutive 1s short clips and study the effect of different clip sampling rates. Sampling clips every 1s achieves the best performance while sampling clips every 8s achieves the best efficiency (\textbf{8x}) with only \textbf{2.0\%} accuracy drop. This suggests that we can effectively control the accuracy-efficiency trade-off of our framework by varying the clip sampling rate.}
\vspace{-0.1in}
\label{tab:fps}
\end{table}

\subsubsection{Video Processing Configurations}
Clip length and sample rate are important hyper-parameters for sampling short video clips from long video inputs for visual captioning. In this section, we explore the influence of clip length and clip sampling rate on our framework.

\noindent \textbf{Clip Length.}
In Table~\ref{tab:clip_length}, we explore how LVQA performance is influenced by different clip length. We divide the long video into consecutive clips of different length and report the corresponding LVQA accuracy. From the table, we can see that our framework achieves the best accuracy when the clip length is the shortest. As the clip length increases, the performance drops. This suggests that dividing long videos into consecutive 1s short clips is the most effective strategy.

\noindent \textbf{Clip Sampling Rate.}
In Table~\ref{tab:fps}, we explore how LVQA performance is influenced by different clip sampling rate on EgoSchema. Specifically, we divide the input long video into consecutive 1s short clips and change the clip sampling rate to see how LVQA performance changes accordingly. From the table, we can see that sampling one clip every 1s leads to the highest accuracy. Sampling one clip every 8s (i.e., the clip sampling rate of 1/8) achieves \textbf{8x} efficiency while the accuracy drops by only \textbf{2.0\%}. This indicates that we can effectively control the accuracy and efficiency tradeoff of our method by sampling video clips more sparsely.

\subsubsection{LLM Prompt Analysis}
In this section, we (1) analyze several variants of our summarization-based prompt (described in Section~\ref{sec:method}), and (2) experiment with other commonly used prompt designs, including Zero-shot Chain-of-Thought (Zero-shot CoT)~\cite{wei2022chain} and Plan-and-Solve~\cite{wang2023plan}.
Below, we present a detailed analysis of these results.

\noindent\textbf{Multi-round Summarization Prompt.} Given a concatenated set of captions $C$, an input question $Q$, and a set of candidate answers $A$, we can use several input combinations to obtain the summary $S$. Thus, here, we investigate three distinct variants of obtaining summaries $S$:

\begin{itemize}
    \item (C) $\rightarrow$ S: the LLM uses caption-only inputs $C$ to obtain summaries $S$ in the first round of prompting.
    \item (C, Q) $\rightarrow$ S: the LLM uses captions $C$ and a question $Q$ as inputs for generating summaries $S$. Having additional question inputs is beneficial as it allows the LLM to generate a summary $S$ specifically tailored for answering an input question $Q$.
    \item (C, Q, A) $\rightarrow$ S: the LLM takes captions $C$, a question $Q$, and the answer candidates $A$ as its inputs to produce summaries $S$. Having additional answer candidate inputs enables the LLM to generate a summary $S$ most tailored to particular question-answer pairs.
\end{itemize}

\begin{table}[t]
\centering
\resizebox{\columnwidth}{!}{
    \begin{tabular}{lcccc}
    \toprule
    Prompt Type & Standard & (C) $\rightarrow$ S & (C, Q) $\rightarrow$ S & (C, Q, A) $\rightarrow$ S  \\ 
    \midrule
    Acc. (\%) & 55.2 & 55.0 & \textbf{58.8} & 54.8  \\
    \bottomrule
    \end{tabular}
}
\caption{\textbf{Different variants of our multi-round summarization prompt.} Our results indicate that the (C, Q) $\rightarrow$ S variant that takes concatenated captions $C$ and a question $Q$ for generating a summary $S$ works the best, significantly outperforming (\textbf{+3.6\%}) the standard prompt. This confirms our hypothesis that additional inputs in the form of a question $Q$ enable the LLM to generate a summary $S$ tailored to a given question $Q$.}
\label{tab:variants}
\end{table}

In Table~\ref{tab:variants}, we explore the effectiveness of these three prompt variants. We observe that while the (C) $\rightarrow$ S and the (C, Q, A) variant $\rightarrow$ S perform similarly to the standard baseline, the (C, Q) $\rightarrow$ S variant greatly outperforms the standard baseline by \textbf{3.6\%}. Compared with (C) $\rightarrow$ S, (C, Q) $\rightarrow$ S incorporates a given question as the input and thus leads to a big boost in LVQA performance. This confirms our earlier intuition that having additional question $Q$ inputs enables the LLM to generate a summary $S$ specifically tailored for answering that question. However, adding answer candidates $A$ as additional inputs (i.e., the (C, Q, A) $\rightarrow$ S variant) leads to a drop in performance (\textbf{-4.0\%}) compared with the (C, Q) $\rightarrow$ S variant. We conjecture that this might happen because the candidate answers $A$ in EgoSchema are often very long, and thus, they may mislead/distract the LLM into generating a suboptimal summary $S$.

\begin{table}[t]
\centering
\setlength{\tabcolsep}{20pt}
\resizebox{\columnwidth}{!}{
\begin{tabular}{lc}
\toprule
Prompting Technique & Acc. (\%)\\ 
\midrule
Standard & 55.2 \\ 
Plan-and-Solve~\cite{wang2023plan} & 55.2 \\
Chain-of-Thought~\cite{wei2022chain} & 57.8 \\
Ours & \textbf{58.8} \\
\bottomrule
\end{tabular}
}
\caption{\textbf{Comparison with commonly used prompting techniques.} The ``Standard" means a standard LVQA prompt (see Section~\ref{sec:method}). Our multi-round summarization prompt performs best.}
\vspace{-0.1in}
\label{tab:prompt}
\end{table}

\noindent\textbf{Comparison with Commonly Used Prompts.} Next, in Table~\ref{tab:prompt}, we compare our multi-round summarization prompt with other commonly used prompts such as Zero-shot Chain-of-Thought~\cite{wei2022chain} and Plan-and-Solve~\cite{wang2023plan}. Our results indicate that our multi-round summarization prompt achieves the best performance among all of these prompts. Furthermore, we note that it outperforms the standard prompt (described in Section~\ref{sec:method}) by a substantial \textbf{3.6\%} in LVQA accuracy, thus indicating the effectiveness of our prompt design. 

\noindent\textbf{Efficiency Analysis.}
We compare the efficiency of our multi-round summarization prompt and the standard prompt within our entire framework. We report that for a 3-minute EgoSchema video, the LaViLa captioner takes \textbf{22.36s} to generate all short-term captions on a single A6000 GPU. The standard prompt using GPT-3.5 as the LLM then takes \textbf{0.4s} for processing the captions from the 3-minute video, while the multi-round summarization prompt takes \textbf{3.6s}. Therefore, the additional computational cost introduced by the multi-round summarization prompt is relatively small compared to the total runtime, which shows the efficiency of our multi-round summarization prompt. We also note that such a small increase in runtime leads to a substantial \textbf{9.4\%} increase in QA accuracy on the full set of EgoSchema compared to using the standard prompt as shown in Table~\ref{tab:egoschema}.

\begin{table}[t]
    \centering
    \setlength{\tabcolsep}{3pt}
    \resizebox{0.95\columnwidth}{!}{
    \begin{tabular}{lcccc}
    \toprule
    \multirow{2}{*}{Method} & \multirow{2}{*}{LM} & \multirow{2}{*}{Params} & Throughput & \multirow{2}{*}{Acc. (\%)} \\ 
     &  &  & (video / s) &  \\
    \midrule
    FrozenBiLM & DeBERTa & 900M & - & 26.9 \\
    mPLUG-Owl & LLaMA & 7B & - & 31.1 \\
    InternVideo & Transformer & 478M & - & 32.1 \\
    LongViViT & BERT & 1B & - & 33.3 \\
    Video ChatCaptioner & GPT-3.5 & 175B & 1.24 & 39.0 \\
    VLog & GPT-3.5 & 175B & 1.04 & 44.0 \\
    Vamos & GPT-4 & 1.5T & - & 48.3 \\
    \midrule
    \model~(Ours) & \multirow{2}{*}{GPT-3.5} & \multirow{2}{*}{175B} & \multirow{2}{*}{2.63} & \multirow{2}{*}{42.8} \\
    \textit{w/ Standard Prompt} & & &  \\
    \model~(Ours) & \multirow{2}{*}{GPT-3.5} & \multirow{2}{*}{175B} & \multirow{2}{*}{2.31} & \multirow{2}{*}{\textbf{52.2}} \\
    \textit{w/ Summarization Prompt} & & &  \\
    \bottomrule
    \end{tabular}
    }
    \caption{\textbf{Main results on the full set of EgoSchema.}
    The throughput is measured by the number of 3-minute videos that a method can process in one minute using an A6000 GPU.
    Our \model~framework with the proposed multi-round summarization prompt achieves \textbf{52.2\%} accuracy, outperforming the variant of our model with a standard prompt by a significant margin (\textbf{9.4\%}). Additionally, our method outperforms the previous best-performing Vamos model by \textbf{3.9\%} despite using a weaker LLM, as well as all other competing methods. Our method also has the highest throughput compared with other LLM-based methods.}
    \label{tab:egoschema}
    \vspace{-0.1in}
\end{table}

\subsection{Main Results on EgoSchema} \label{main_result}
In Table~\ref{tab:egoschema}, we evaluate our best-performing \model~framework on the full EgoSchema test set containing 5K video samples. We compare our approach with prior state-of-the-art methods including InternVideo~\cite{wang2022internvideo}, mPLUG-Owl~\cite{ye2023mplug}, FrozenBiLM~\cite{yang2022frozenbilm}, Video ChatCaptioner~\cite{chen2023video}, VLog~\cite{vlog}, as well as the concurrent works of LongViViT~\cite{papalampidi2023simple}, and Vamos~\cite{wang2023vamos}. The throughput is measured by the number of 3-minute videos that a method can process in one minute using an A6000 GPU.

Based on these results, we observe that our \model~framework with the proposed multi-round summarization prompt achieves \textbf{52.2\%} accuracy, outperforming the concurrent Vamos model by \textbf{+3.9\%} despite using a weaker LLM (GPT-3.5) than their approach (GPT-4). We also observe that our model outperforms all other baselines by an even more significant margin ($>$\textbf{8.2\%}). Additionally, we can see that our method has the highest throughput compared with other LLM-based approaches. This shows that our framework is the most efficient while achieving the highest accuracy. Lastly, our results indicate that using our novel multi-round summarization prompt outperforms the variant of our approach with the standard prompt by a significant margin of \textbf{9.4\%}. These results validate the effectiveness of our LLM-based framework design.

\subsection{Results on Other Datasets}
\begin{table}[t]
\centering
\setlength{\tabcolsep}{3pt}
\resizebox{\columnwidth}{!}{
\begin{tabular}{lcccccc}
\toprule
Method & LM & Params & C. & T. & D. & All  \\ 
\midrule
VFC & Transformer & 164M & 45.4 & 51.6 & 64.1 & 51.5 \\
InternVideo & Transformer & 478M & 43.4 & 48.0 & 65.1 & 49.1 \\
ViperGPT & GPT-3 & 175B & - & -  & - & 60.0 \\
SeViLA & Flan-T5 & 4B & 61.3 & 61.5 & 75.6 & 63.6 \\
\midrule
\model~(ours) & GPT-3.5 & 175B & 67.1 & 60.1 & 76.5 & 66.3\\
\model~(ours) & GPT-4 & 1.8T & \textbf{73.7} & \textbf{70.2} & \textbf{81.9} & \textbf{73.8} \\
\bottomrule
\end{tabular}
}
\caption{\textbf{Zero-shot results on NExT-QA.} C, T, D is short for Causual, Temporal, Descriptive, respectively. The best variant of LLoVi achieves \textbf{73.8\%} accuracy, outperforming previous best-performing model SeViLA by \textbf{10.2\%}. 
}
\label{tab:nextqa}
\end{table}

\begin{table}[t]
\centering
\setlength{\tabcolsep}{10pt}
\resizebox{\columnwidth}{!}{
\begin{tabular}{lccc}
\toprule
Method & LM & Params & Acc. (\%)  \\ 
\midrule
\multicolumn{2}{l}{\textit{Supervised}} \\
HQGA & - & 46M & 47.7 \\
VGT & Transformer & 511M & 51.3 \\
BlindGPT & GPT-3 & 175B & 51.6 \\
CaVIR & GPT-3 & 175B & 57.6 \\
\midrule
\multicolumn{2}{l}{\textit{Zero-shot}} \\
SeViLA & Flan-T5 & 4B &  60.9 \\
\model~(ours) & GPT-4 & 1.8T & \textbf{67.1} \\
\bottomrule
\end{tabular}
}
\caption{\textbf{Results on IntentQA.} Our zero-shot framework outperforms previous supervised methods by a large margin (\textbf{9.5\%}). LLoVi also outperforms the recent state-of-the-art zero-shot method, SeViLA, by \textbf{6.2\%}. 
}
\label{tab:intentqa}
\vspace{-0.1in}
\end{table}

\begin{table}[t]
\centering
\setlength{\tabcolsep}{1pt}
\resizebox{\columnwidth}{!}{
\begin{tabular}{lcccccccccc}
\toprule
\multirow{2}{*}{Method} & \multirow{2}{*}{LM} & \multirow{2}{*}{Params} & mIoP & IoP & mIoU & IoU5 & Acc \\ 
& & & & @0.5 & & @0.5 & @ GQA \\
\midrule
\multicolumn{6}{l}{\textit{Weakly-Supervised}} \\
IGV & - & 110M & 21.4 & 18.9 & 14.0 & 9.6 & 10.2 \\
Temp[CLIP] & Transformer & 130M & 25.7 & 25.5 & 12.1 & 8.9 & 16.0 \\
FrozenBiLM & DeBERTa & 900M & 24.2 & 23.7 & 9.6 & 6.1 & 17.5 \\
\color{gray}SeViLA & \color{gray}Flan-T5 & \color{gray}4B & \color{gray}29.5 & \color{gray}22.9 & \color{gray}21.7 & \color{gray}13.8 & \color{gray}16.6\\
\midrule
\multicolumn{6}{l}{\textit{Zero-shot}} \\
\model~(ours) & GPT-4 & 1.8T & \textbf{39.4} & \textbf{38.0} & \textbf{21.5} & \textbf{16.2} & \textbf{26.8} \\
\bottomrule
\end{tabular}
}
\caption{\textbf{Grounded LVQA results on NExT-GQA.} We extend LLoVi to the grounded LVQA task and show that it outperforms prior weakly-supervised approaches on all evaluation metrics. For a fair comparison, we \text{\color{gray}de-emphasize} the models that were pretrained using video-language grounding annotations.}
\label{tab:nextgqa}
\end{table}

Next, we demonstrate that our \model~framework generalizes well to other LVQA benchmarks.

\noindent
\textbf{NExT-QA.} In Table~\ref{tab:nextqa}, we evaluate LLoVi on the NExT-QA~\cite{xiao2021next} validation set in a zero-shot setting. We compare our approach with prior methods: VFC~\cite{momeni2023verbs}, InternVideo~\cite{wang2022internvideo}, ViperGPT~\cite{suris2023vipergpt}, and SeViLA~\cite{yu2023self}. We observe that the best variant of LLoVi outperforms the previous best-performing method, SeViLA by a significant margin of \textbf{10.2\%}. 
We conjecture this improvement comes from our decomposition of LVQA into two stages, i.e., short-term captioning followed by long-term reasoning with an LLM, which enables us to harness the power of modern LLMs for this challenging task. 

\noindent
\textbf{IntentQA.} In Table~\ref{tab:intentqa}, we evaluate our method on the IntentQA~\cite{li2023intentqa} test set. In our comparisons, we include several fully supervised methods (HQGA~\cite{xiao2021video}, VGT~\cite{xiao2022video}, BlindGPT~\cite{ouyang2022training}, CaVIR~\cite{Li_2023_ICCV}) and the recent state-of-the-art zero-shot approach, SeViLA. From the results in Table~\ref{tab:intentqa}, we observe that our method greatly outperforms all prior approaches. 

\noindent
\textbf{NExT-GQA.} In Table~\ref{tab:nextgqa}, we extend our framework to the grounded LVQA task and evaluate it on the NExT-GQA~\cite{xiao2023can} test set. To do this, we extract visual captions from each frame and then provide them, along with their corresponding frame indices, to the LLM to identify the required frame indices for answering the question. More details are provided in the Supplementary Material. We compare LLoVi with the weakly-supervised methods: IGV~\cite{li2022invariant}, Temp[CLIP](NG+)~\cite{xiao2023can}, FrozenBiLM (NG+)~\cite{xiao2023can} and SeViLA~\cite{yu2023self}. These baselines are first trained on NExT-GQA to maximize the QA accuracy and then use ad-hoc methods~\cite{xiao2023can} to estimate a relevant video segment for question-answering. Although LLoVi is not trained on NExT-GQA, it still outperforms these weakly-supervised methods by a large margin according to all evaluation metrics. These results demonstrate that our framework can be used to temporally ground its predictions for more explainable long-range video understanding.

\section{Conclusion}

In this work, we present a simple, yet highly effective LLM-based framework for long-range video question-answering (LVQA). Our framework outperforms all prior models on the newly introduced EgoSchema benchmark. 
Furthermore, we demonstrate that our approach generalizes to other LVQA benchmarks such as NExT-QA, IntentQA, and it can also be extended to grounded LVQA tasks. 
Lastly, we thoroughly evaluate various design choices of our approach and analyze the key factors behind the success of our method. We hope that our simple LVQA framework will help inspire new ideas and simplify model design in long-range video understanding.

\section*{Limitations}

One limitation of our approach is that it might produce suboptimal results if the visual captioning outputs are inaccurate. This might happen because many existing visual captioners suffer from hallucinations and often struggle to effectively capture fine-grained visual details (e.g., fine-grained human-object interactions, etc.). Having said this, our framework is highly flexible and agnostic to the exact visual captioning model that it uses. Thus, we believe that in the future, we will be able to address this limitation by leveraging more powerful visual captioners. Furthermore, another limitation of our approach is that many modern LLMs are not designed for long-context modeling, which is critical for the LVQA task. However, we believe that this limitation will also be addressed in the future via a more sophisticated LLM design, thus, allowing us to incorporate more powerful LLMs for even better LVQA performance.

\section*{Acknowledgements}
We thank Karttikeya Mangalam, Feng Cheng, Yan-Bo Lin, Yue Yang, and Soumitri Chattopadhyay for their discussion and valuable feedback. This work was supported by the Sony Faculty Innovation Award, Laboratory for Analytic Sciences via NC State University, and ONR Award N00014-23-1-2356.


\bibliography{anthology,main}

\begin{thebibliography}{94}
\expandafter\ifx\csname natexlab\endcsname\relax\def\natexlab#1{#1}\fi

\bibitem[{Bertasius et~al.(2021)Bertasius, Wang, and Torresani}]{bertasius2021space}
Gedas Bertasius, Heng Wang, and Lorenzo Torresani. 2021.
\newblock Is space-time attention all you need for video understanding?
\newblock In \emph{ICML}, volume~2, page~4.

\bibitem[{Bhattacharya et~al.(2023)Bhattacharya, Singla, Krishnamurthy, Shah, and Chen}]{bhattacharya2023video}
Aanisha Bhattacharya, Yaman~K Singla, Balaji Krishnamurthy, Rajiv~Ratn Shah, and Changyou Chen. 2023.
\newblock A video is worth 4096 tokens: Verbalize story videos to understand them in zero shot.
\newblock \emph{arXiv preprint arXiv:2305.09758}.

\bibitem[{Brown et~al.(2020)Brown, Mann, Ryder, Subbiah, Kaplan, Dhariwal, Neelakantan, Shyam, Sastry, Askell et~al.}]{brown2020language}
Tom Brown, Benjamin Mann, Nick Ryder, Melanie Subbiah, Jared~D Kaplan, Prafulla Dhariwal, Arvind Neelakantan, Pranav Shyam, Girish Sastry, Amanda Askell, et~al. 2020.
\newblock Language models are few-shot learners.
\newblock \emph{Advances in neural information processing systems}, 33:1877--1901.

\bibitem[{Chen et~al.(2023)Chen, Zhu, Haydarov, Li, and Elhoseiny}]{chen2023video}
Jun Chen, Deyao Zhu, Kilichbek Haydarov, Xiang Li, and Mohamed Elhoseiny. 2023.
\newblock Video chatcaptioner: Towards the enriched spatiotemporal descriptions.
\newblock \emph{arXiv preprint arXiv:2304.04227}.

\bibitem[{Chen et~al.(2022)Chen, Ma, Wang, and Cohen}]{chen2022program}
Wenhu Chen, Xueguang Ma, Xinyi Wang, and William~W Cohen. 2022.
\newblock Program of thoughts prompting: Disentangling computation from reasoning for numerical reasoning tasks.
\newblock \emph{arXiv preprint arXiv:2211.12588}.

\bibitem[{Chen et~al.(2020)Chen, Cao, Hu, and Wang}]{chen2020memory}
Yihong Chen, Yue Cao, Han Hu, and Liwei Wang. 2020.
\newblock Memory enhanced global-local aggregation for video object detection.
\newblock In \emph{Proceedings of the IEEE/CVF conference on computer vision and pattern recognition}, pages 10337--10346.

\bibitem[{Cheng and Bertasius(2022)}]{cheng2022tallformer}
Feng Cheng and Gedas Bertasius. 2022.
\newblock Tallformer: Temporal action localization with a long-memory transformer.
\newblock In \emph{European Conference on Computer Vision}, pages 503--521. Springer.

\bibitem[{Cheng et~al.(2023)Cheng, Wang, Lei, Crandall, Bansal, and Bertasius}]{VindLU_CVPR2023}
Feng Cheng, Xizi Wang, Jie Lei, David Crandall, Mohit Bansal, and Gedas Bertasius. 2023.
\newblock Vindlu: A recipe for effective video-and-language pretraining.
\newblock In \emph{The IEEE Conference on Computer Vision and Pattern Recognition (CVPR)}.

\bibitem[{Choi et~al.(2021)Choi, On, Heo, Seo, Jang, Lee, and Zhang}]{DBLP:conf/aaai/ChoiOHSJLZ21}
Seongho Choi, Kyoung{-}Woon On, Yu{-}Jung Heo, Ahjeong Seo, Youwon Jang, Min~Su Lee, and Byoung{-}Tak Zhang. 2021.
\newblock \href {https://doi.org/10.1609/AAAI.V35I2.16203} {Dramaqa: Character-centered video story understanding with hierarchical {QA}}.
\newblock In \emph{Thirty-Fifth {AAAI} Conference on Artificial Intelligence, {AAAI} 2021, Thirty-Third Conference on Innovative Applications of Artificial Intelligence, {IAAI} 2021, The Eleventh Symposium on Educational Advances in Artificial Intelligence, {EAAI} 2021, Virtual Event, February 2-9, 2021}, pages 1166--1174. {AAAI} Press.

\bibitem[{Chung et~al.(2022)Chung, Hou, Longpre, Zoph, Tay, Fedus, Li, Wang, Dehghani, Brahma et~al.}]{chung2022scaling}
Hyung~Won Chung, Le~Hou, Shayne Longpre, Barret Zoph, Yi~Tay, William Fedus, Yunxuan Li, Xuezhi Wang, Mostafa Dehghani, Siddhartha Brahma, et~al. 2022.
\newblock Scaling instruction-finetuned language models.
\newblock \emph{arXiv preprint arXiv:2210.11416}.

\bibitem[{Chung and Yu(2023)}]{chung2023long}
Jiwan Chung and Youngjae Yu. 2023.
\newblock Long story short: a summarize-then-search method for long video question answering.
\newblock In \emph{BMVC}.

\bibitem[{Fu et~al.(2021)Fu, Li, Gan, Lin, Wang, Wang, and Liu}]{fu2021violet}
Tsu-Jui Fu, Linjie Li, Zhe Gan, Kevin Lin, William~Yang Wang, Lijuan Wang, and Zicheng Liu. 2021.
\newblock {VIOLET: End-to-End Video-Language Transformers with Masked Visual-token Modeling}.
\newblock In \emph{arXiv:2111.1268}.

\bibitem[{Grauman et~al.(2022)Grauman, Westbury, Byrne, Chavis, Furnari, Girdhar, Hamburger, Jiang, Liu, Liu et~al.}]{grauman2022ego4d}
Kristen Grauman, Andrew Westbury, Eugene Byrne, Zachary Chavis, Antonino Furnari, Rohit Girdhar, Jackson Hamburger, Hao Jiang, Miao Liu, Xingyu Liu, et~al. 2022.
\newblock Ego4d: Around the world in 3,000 hours of egocentric video.
\newblock In \emph{Proceedings of the IEEE/CVF Conference on Computer Vision and Pattern Recognition}, pages 18995--19012.

\bibitem[{Gu et~al.(2021)Gu, Goel, and R{\'e}}]{gu2021efficiently}
Albert Gu, Karan Goel, and Christopher R{\'e}. 2021.
\newblock Efficiently modeling long sequences with structured state spaces.
\newblock \emph{arXiv preprint arXiv:2111.00396}.

\bibitem[{Gur et~al.(2023)Gur, Furuta, Huang, Safdari, Matsuo, Eck, and Faust}]{gur2023real}
Izzeddin Gur, Hiroki Furuta, Austin Huang, Mustafa Safdari, Yutaka Matsuo, Douglas Eck, and Aleksandra Faust. 2023.
\newblock A real-world webagent with planning, long context understanding, and program synthesis.
\newblock \emph{arXiv preprint arXiv:2307.12856}.

\bibitem[{Hao et~al.(2023)Hao, Gu, Ma, Hong, Wang, Wang, and Hu}]{hao2023reasoning}
Shibo Hao, Yi~Gu, Haodi Ma, Joshua~Jiahua Hong, Zhen Wang, Daisy~Zhe Wang, and Zhiting Hu. 2023.
\newblock Reasoning with language model is planning with world model.
\newblock \emph{arXiv preprint arXiv:2305.14992}.

\bibitem[{Holtzman et~al.(2019)Holtzman, Buys, Du, Forbes, and Choi}]{holtzman2019curious}
Ari Holtzman, Jan Buys, Li~Du, Maxwell Forbes, and Yejin Choi. 2019.
\newblock The curious case of neural text degeneration.
\newblock \emph{arXiv preprint arXiv:1904.09751}.

\bibitem[{Hong et~al.(2024)Hong, Wang, Lv, Xu, Yu, Ji, Wang, Wang, Dong, Ding et~al.}]{hong2024cogagent}
Wenyi Hong, Weihan Wang, Qingsong Lv, Jiazheng Xu, Wenmeng Yu, Junhui Ji, Yan Wang, Zihan Wang, Yuxiao Dong, Ming Ding, et~al. 2024.
\newblock Cogagent: A visual language model for gui agents.
\newblock In \emph{Proceedings of the IEEE/CVF Conference on Computer Vision and Pattern Recognition}, pages 14281--14290.

\bibitem[{Hussein et~al.(2019{\natexlab{a}})Hussein, Gavves, and Smeulders}]{hussein2019timeception}
Noureldien Hussein, Efstratios Gavves, and Arnold~WM Smeulders. 2019{\natexlab{a}}.
\newblock Timeception for complex action recognition.
\newblock In \emph{Proceedings of the IEEE/CVF Conference on Computer Vision and Pattern Recognition}, pages 254--263.

\bibitem[{Hussein et~al.(2019{\natexlab{b}})Hussein, Gavves, and Smeulders}]{hussein2019videograph}
Noureldien Hussein, Efstratios Gavves, and Arnold~WM Smeulders. 2019{\natexlab{b}}.
\newblock Videograph: Recognizing minutes-long human activities in videos.
\newblock \emph{arXiv preprint arXiv:1905.05143}.

\bibitem[{Islam and Bertasius(2022)}]{islam2022long}
Md~Mohaiminul Islam and Gedas Bertasius. 2022.
\newblock Long movie clip classification with state-space video models.
\newblock In \emph{European Conference on Computer Vision}, pages 87--104. Springer.

\bibitem[{Islam et~al.(2023)Islam, Hasan, Athrey, Braskich, and Bertasius}]{islam2023efficient}
Md~Mohaiminul Islam, Mahmudul Hasan, Kishan~Shamsundar Athrey, Tony Braskich, and Gedas Bertasius. 2023.
\newblock Efficient movie scene detection using state-space transformers.
\newblock In \emph{Proceedings of the IEEE/CVF Conference on Computer Vision and Pattern Recognition}, pages 18749--18758.

\bibitem[{Jasani et~al.(2019)Jasani, Girdhar, and Ramanan}]{jasani2019we}
Bhavan Jasani, Rohit Girdhar, and Deva Ramanan. 2019.
\newblock Are we asking the right questions in movieqa?
\newblock In \emph{Proceedings of the IEEE/CVF International Conference on Computer Vision Workshops}, pages 0--0.

\bibitem[{Jiang et~al.(2023)Jiang, Sablayrolles, Mensch, Bamford, Chaplot, Casas, Bressand, Lengyel, Lample, Saulnier et~al.}]{jiang2023mistral}
Albert~Q Jiang, Alexandre Sablayrolles, Arthur Mensch, Chris Bamford, Devendra~Singh Chaplot, Diego de~las Casas, Florian Bressand, Gianna Lengyel, Guillaume Lample, Lucile Saulnier, et~al. 2023.
\newblock Mistral 7b.
\newblock \emph{arXiv preprint arXiv:2310.06825}.

\bibitem[{Kingma and Ba(2014)}]{Kingma2014AdamAM}
Diederik~P. Kingma and Jimmy Ba. 2014.
\newblock \href {https://api.semanticscholar.org/CorpusID:6628106} {Adam: A method for stochastic optimization}.
\newblock \emph{CoRR}, abs/1412.6980.

\bibitem[{Kojima et~al.(2022)Kojima, Gu, Reid, Matsuo, and Iwasawa}]{kojima2022large}
Takeshi Kojima, Shixiang~Shane Gu, Machel Reid, Yutaka Matsuo, and Yusuke Iwasawa. 2022.
\newblock Large language models are zero-shot reasoners.
\newblock \emph{Advances in neural information processing systems}, 35:22199--22213.

\bibitem[{Lee et~al.(2018)Lee, Sung, Yu, and Kim}]{lee2018memory}
Sangho Lee, Jinyoung Sung, Youngjae Yu, and Gunhee Kim. 2018.
\newblock A memory network approach for story-based temporal summarization of 360 videos.
\newblock In \emph{Proceedings of the IEEE conference on computer vision and pattern recognition}, pages 1410--1419.

\bibitem[{Lee et~al.(2021)Lee, Kim, Choi, Kim, and Ro}]{lee2021video}
Sangmin Lee, Hak~Gu Kim, Dae~Hwi Choi, Hyung-Il Kim, and Yong~Man Ro. 2021.
\newblock Video prediction recalling long-term motion context via memory alignment learning.
\newblock In \emph{Proceedings of the IEEE/CVF Conference on Computer Vision and Pattern Recognition}, pages 3054--3063.

\bibitem[{Lei et~al.(2021)Lei, Li, Zhou, Gan, Berg, Bansal, and Liu}]{lei2021less}
Jie Lei, Linjie Li, Luowei Zhou, Zhe Gan, Tamara~L. Berg, Mohit Bansal, and Jingjing Liu. 2021.
\newblock Less is more: Clipbert for video-and-language learningvia sparse sampling.
\newblock In \emph{CVPR}.

\bibitem[{Lei et~al.(2018)Lei, Yu, Bansal, and Berg}]{lei2018tvqa}
Jie Lei, Licheng Yu, Mohit Bansal, and Tamara~L Berg. 2018.
\newblock Tvqa: Localized, compositional video question answering.
\newblock In \emph{EMNLP}.

\bibitem[{Li et~al.(2023{\natexlab{a}})Li, Wei, Han, and Fan}]{li2023intentqa}
Jiapeng Li, Ping Wei, Wenjuan Han, and Lifeng Fan. 2023{\natexlab{a}}.
\newblock Intentqa: Context-aware video intent reasoning.
\newblock In \emph{Proceedings of the IEEE/CVF International Conference on Computer Vision}, pages 11963--11974.

\bibitem[{Li et~al.(2023{\natexlab{b}})Li, Wei, Han, and Fan}]{Li_2023_ICCV}
Jiapeng Li, Ping Wei, Wenjuan Han, and Lifeng Fan. 2023{\natexlab{b}}.
\newblock Intentqa: Context-aware video intent reasoning.
\newblock In \emph{Proceedings of the IEEE/CVF International Conference on Computer Vision (ICCV)}, pages 11963--11974.

\bibitem[{Li et~al.(2023{\natexlab{c}})Li, Li, Savarese, and Hoi}]{li2023blip2}
Junnan Li, Dongxu Li, Silvio Savarese, and Steven Hoi. 2023{\natexlab{c}}.
\newblock {BLIP-2:} bootstrapping language-image pre-training with frozen image encoders and large language models.
\newblock In \emph{ICML}.

\bibitem[{Li et~al.(2023{\natexlab{d}})Li, Li, Savarese, and Hoi}]{li2023blip}
Junnan Li, Dongxu Li, Silvio Savarese, and Steven Hoi. 2023{\natexlab{d}}.
\newblock {BLIP-2:} bootstrapping language-image pre-training with frozen image encoders and large language models.
\newblock In \emph{ICML}.

\bibitem[{Li et~al.(2023{\natexlab{e}})Li, He, Wang, Li, Wang, Luo, Wang, Wang, and Qiao}]{li2023videochat}
KunChang Li, Yinan He, Yi~Wang, Yizhuo Li, Wenhai Wang, Ping Luo, Yali Wang, Limin Wang, and Yu~Qiao. 2023{\natexlab{e}}.
\newblock \href {http://arxiv.org/abs/2305.06355} {Videochat: Chat-centric video understanding}.

\bibitem[{Li et~al.(2022)Li, Wang, Xiao, Ji, and Chua}]{li2022invariant}
Yicong Li, Xiang Wang, Junbin Xiao, Wei Ji, and Tat-Seng Chua. 2022.
\newblock Invariant grounding for video question answering.
\newblock In \emph{Proceedings of the IEEE/CVF Conference on Computer Vision and Pattern Recognition}, pages 2928--2937.

\bibitem[{Lin et~al.(2023{\natexlab{a}})Lin, Ahmed, Li, Lin, Azarnasab, Yang, Wang, Liang, Liu, Lu, Liu, and Wang}]{2023mmvid}
Kevin Lin, Faisal Ahmed, Linjie Li, Chung-Ching Lin, Ehsan Azarnasab, Zhengyuan Yang, Jianfeng Wang, Lin Liang, Zicheng Liu, Yumao Lu, Ce~Liu, and Lijuan Wang. 2023{\natexlab{a}}.
\newblock Mm-vid: Advancing video understanding with gpt-4v(ision).
\newblock \emph{arXiv preprint arXiv:2310.19773}.

\bibitem[{Lin et~al.(2023{\natexlab{b}})Lin, Ahmed, Li, Lin, Azarnasab, Yang, Wang, Liang, Liu, Lu et~al.}]{lin2023mm}
Kevin Lin, Faisal Ahmed, Linjie Li, Chung-Ching Lin, Ehsan Azarnasab, Zhengyuan Yang, Jianfeng Wang, Lin Liang, Zicheng Liu, Yumao Lu, et~al. 2023{\natexlab{b}}.
\newblock Mm-vid: Advancing video understanding with gpt-4v (ision).
\newblock \emph{arXiv preprint arXiv:2310.19773}.

\bibitem[{Lin and Lei(2023)}]{vlog}
Kevin~Qinghong Lin and Stan~Weixian Lei. 2023.
\newblock \href {https://github.com/showlab/VLog} {Vlog: Video as a long document}.

\bibitem[{Liu et~al.(2023{\natexlab{a}})Liu, Jiang, Zhang, Liu, Zhang, Biswas, and Stone}]{liu2023llm}
Bo~Liu, Yuqian Jiang, Xiaohan Zhang, Qiang Liu, Shiqi Zhang, Joydeep Biswas, and Peter Stone. 2023{\natexlab{a}}.
\newblock Llm+ p: Empowering large language models with optimal planning proficiency.
\newblock \emph{arXiv preprint arXiv:2304.11477}.

\bibitem[{Liu et~al.(2023{\natexlab{b}})Liu, Li, Wu, and Lee}]{liu2023llava}
Haotian Liu, Chunyuan Li, Qingyang Wu, and Yong~Jae Lee. 2023{\natexlab{b}}.
\newblock Visual instruction tuning.
\newblock In \emph{NeurIPS}.

\bibitem[{Mangalam et~al.(2023)Mangalam, Akshulakov, and Malik}]{mangalam2023egoschema}
Karttikeya Mangalam, Raiymbek Akshulakov, and Jitendra Malik. 2023.
\newblock Egoschema: A diagnostic benchmark for very long-form video language understanding.
\newblock \emph{arXiv preprint arXiv:2308.09126}.

\bibitem[{Momeni et~al.(2023)Momeni, Caron, Nagrani, Zisserman, and Schmid}]{momeni2023verbs}
Liliane Momeni, Mathilde Caron, Arsha Nagrani, Andrew Zisserman, and Cordelia Schmid. 2023.
\newblock Verbs in action: Improving verb understanding in video-language models.
\newblock In \emph{Proceedings of the IEEE/CVF International Conference on Computer Vision}, pages 15579--15591.

\bibitem[{Nguyen et~al.(2022)Nguyen, Goel, Gu, Downs, Shah, Dao, Baccus, and R{\'e}}]{nguyen2022s4nd}
Eric Nguyen, Karan Goel, Albert Gu, Gordon Downs, Preey Shah, Tri Dao, Stephen Baccus, and Christopher R{\'e}. 2022.
\newblock S4nd: Modeling images and videos as multidimensional signals with state spaces.
\newblock \emph{Advances in neural information processing systems}, 35:2846--2861.

\bibitem[{OpenAI(2023{\natexlab{a}})}]{gpt35}
OpenAI. 2023{\natexlab{a}}.
\newblock \href {https://platform.openai.com/docs/models/gpt-3-5-turbo} {Gpt-3.5}.

\bibitem[{OpenAI(2023{\natexlab{b}})}]{openai2023gpt4}
OpenAI. 2023{\natexlab{b}}.
\newblock \href {http://arxiv.org/abs/2303.08774} {Gpt-4 technical report}.

\bibitem[{Ouyang et~al.(2022)Ouyang, Wu, Jiang, Almeida, Wainwright, Mishkin, Zhang, Agarwal, Slama, Ray et~al.}]{ouyang2022training}
Long Ouyang, Jeffrey Wu, Xu~Jiang, Diogo Almeida, Carroll Wainwright, Pamela Mishkin, Chong Zhang, Sandhini Agarwal, Katarina Slama, Alex Ray, et~al. 2022.
\newblock Training language models to follow instructions with human feedback.
\newblock \emph{Advances in Neural Information Processing Systems}, 35:27730--27744.

\bibitem[{Papalampidi et~al.(2023)Papalampidi, Koppula, Pathak, Chiu, Heyward, Patraucean, Shen, Miech, Zisserman, and Nematzdeh}]{papalampidi2023simple}
Pinelopi Papalampidi, Skanda Koppula, Shreya Pathak, Justin Chiu, Joe Heyward, Viorica Patraucean, Jiajun Shen, Antoine Miech, Andrew Zisserman, and Aida Nematzdeh. 2023.
\newblock A simple recipe for contrastively pre-training video-first encoders beyond 16 frames.
\newblock \emph{arXiv preprint arXiv:2312.07395}.

\bibitem[{Pryzant et~al.(2023)Pryzant, Iter, Li, Lee, Zhu, and Zeng}]{pryzant2023automatic}
Reid Pryzant, Dan Iter, Jerry Li, Yin~Tat Lee, Chenguang Zhu, and Michael Zeng. 2023.
\newblock Automatic prompt optimization with" gradient descent" and beam search.
\newblock \emph{arXiv preprint arXiv:2305.03495}.

\bibitem[{Qinghong~Lin et~al.(2022)Qinghong~Lin, Jinpeng~Wang, Soldan, Wray, Yan, Zhongcong~Xu, Gao, Tu, Zhao, Kong et~al.}]{qinghong2022egocentric}
Kevin Qinghong~Lin, Alex Jinpeng~Wang, Mattia Soldan, Michael Wray, Rui Yan, Eric Zhongcong~Xu, Difei Gao, Rongcheng Tu, Wenzhe Zhao, Weijie Kong, et~al. 2022.
\newblock Egocentric video-language pretraining.
\newblock \emph{arXiv e-prints}, pages arXiv--2206.

\bibitem[{Radford et~al.(2021)Radford, Kim, Hallacy, Ramesh, Goh, Agarwal, Sastry, Askell, Mishkin, Clark et~al.}]{radford2021learning}
Alec Radford, Jong~Wook Kim, Chris Hallacy, Aditya Ramesh, Gabriel Goh, Sandhini Agarwal, Girish Sastry, Amanda Askell, Pamela Mishkin, Jack Clark, et~al. 2021.
\newblock Learning transferable visual models from natural language supervision.
\newblock In \emph{International conference on machine learning}, pages 8748--8763. PMLR.

\bibitem[{Radford et~al.(2019)Radford, Wu, Child, Luan, Amodei, Sutskever et~al.}]{radford2019language}
Alec Radford, Jeffrey Wu, Rewon Child, David Luan, Dario Amodei, Ilya Sutskever, et~al. 2019.
\newblock Language models are unsupervised multitask learners.
\newblock \emph{OpenAI blog}, 1(8):9.

\bibitem[{Raffel et~al.(2020)Raffel, Shazeer, Roberts, Lee, Narang, Matena, Zhou, Li, and Liu}]{raffel2020exploring}
Colin Raffel, Noam Shazeer, Adam Roberts, Katherine Lee, Sharan Narang, Michael Matena, Yanqi Zhou, Wei Li, and Peter~J Liu. 2020.
\newblock Exploring the limits of transfer learning with a unified text-to-text transformer.
\newblock \emph{The Journal of Machine Learning Research}, 21(1):5485--5551.

\bibitem[{Schick and Sch{\"u}tze(2020)}]{schick2020exploiting}
Timo Schick and Hinrich Sch{\"u}tze. 2020.
\newblock Exploiting cloze questions for few shot text classification and natural language inference.
\newblock \emph{arXiv preprint arXiv:2001.07676}.

\bibitem[{Song et~al.(2023{\natexlab{a}})Song, Wu, Washington, Sadler, Chao, and Su}]{song2023llm}
Chan~Hee Song, Jiaman Wu, Clayton Washington, Brian~M Sadler, Wei-Lun Chao, and Yu~Su. 2023{\natexlab{a}}.
\newblock Llm-planner: Few-shot grounded planning for embodied agents with large language models.
\newblock In \emph{Proceedings of the IEEE/CVF International Conference on Computer Vision}, pages 2998--3009.

\bibitem[{Song et~al.(2023{\natexlab{b}})Song, Chai, Wang, Zhang, Zhou, Wu, Guo, Ye, Lu, Hwang et~al.}]{song2023moviechat}
Enxin Song, Wenhao Chai, Guanhong Wang, Yucheng Zhang, Haoyang Zhou, Feiyang Wu, Xun Guo, Tian Ye, Yan Lu, Jenq-Neng Hwang, et~al. 2023{\natexlab{b}}.
\newblock Moviechat: From dense token to sparse memory for long video understanding.
\newblock \emph{arXiv preprint arXiv:2307.16449}.

\bibitem[{Sun et~al.(2023)Sun, Liu, Wang, Zhu, and Iyyer}]{sun2023pearl}
Simeng Sun, Yang Liu, Shuohang Wang, Chenguang Zhu, and Mohit Iyyer. 2023.
\newblock Pearl: Prompting large language models to plan and execute actions over long documents.
\newblock \emph{arXiv preprint arXiv:2305.14564}.

\bibitem[{Sun et~al.(2022)Sun, Xue, Song, Liu, Yang, and Fu}]{sun2022long}
Yuchong Sun, Hongwei Xue, Ruihua Song, Bei Liu, Huan Yang, and Jianlong Fu. 2022.
\newblock Long-form video-language pre-training with multimodal temporal contrastive learning.
\newblock \emph{Advances in neural information processing systems}, 35:38032--38045.

\bibitem[{Sur\'is et~al.(2023)Sur\'is, Menon, and Vondrick}]{suris2023vipergpt}
D\'idac Sur\'is, Sachit Menon, and Carl Vondrick. 2023.
\newblock Vipergpt: Visual inference via python execution for reasoning.
\newblock \emph{Proceedings of IEEE International Conference on Computer Vision (ICCV)}.

\bibitem[{Tapaswi et~al.(2016)Tapaswi, Zhu, Stiefelhagen, Torralba, Urtasun, and Fidler}]{tapaswi2016movieqa}
Makarand Tapaswi, Yukun Zhu, Rainer Stiefelhagen, Antonio Torralba, Raquel Urtasun, and Sanja Fidler. 2016.
\newblock Movieqa: Understanding stories in movies through question-answering.
\newblock In \emph{Proceedings of the IEEE conference on computer vision and pattern recognition}, pages 4631--4640.

\bibitem[{Tay et~al.(2022)Tay, Dehghani, Tran, Garcia, Wei, Wang, Chung, Bahri, Schuster, Zheng et~al.}]{tay2022ul2}
Yi~Tay, Mostafa Dehghani, Vinh~Q Tran, Xavier Garcia, Jason Wei, Xuezhi Wang, Hyung~Won Chung, Dara Bahri, Tal Schuster, Steven Zheng, et~al. 2022.
\newblock Ul2: Unifying language learning paradigms.
\newblock In \emph{The Eleventh International Conference on Learning Representations}.

\bibitem[{Touvron et~al.(2023)Touvron, Martin, Stone, Albert, Almahairi, Babaei, Bashlykov, Batra, Bhargava, Bhosale et~al.}]{touvron2023llama}
Hugo Touvron, Louis Martin, Kevin Stone, Peter Albert, Amjad Almahairi, Yasmine Babaei, Nikolay Bashlykov, Soumya Batra, Prajjwal Bhargava, Shruti Bhosale, et~al. 2023.
\newblock Llama 2: Open foundation and fine-tuned chat models.
\newblock \emph{arXiv preprint arXiv:2307.09288}.

\bibitem[{Wang et~al.(2023{\natexlab{a}})Wang, Zhu, Wang, Yu, Liu, Omar, and Hamid}]{wang2023selective}
Jue Wang, Wentao Zhu, Pichao Wang, Xiang Yu, Linda Liu, Mohamed Omar, and Raffay Hamid. 2023{\natexlab{a}}.
\newblock Selective structured state-spaces for long-form video understanding.
\newblock In \emph{Proceedings of the IEEE/CVF Conference on Computer Vision and Pattern Recognition}, pages 6387--6397.

\bibitem[{Wang et~al.(2023{\natexlab{b}})Wang, Chen, Luo, Dai, Yuan, Wu, and Jiang}]{wang2023chatvideo}
Junke Wang, Dongdong Chen, Chong Luo, Xiyang Dai, Lu~Yuan, Zuxuan Wu, and Yu-Gang Jiang. 2023{\natexlab{b}}.
\newblock Chatvideo: A tracklet-centric multimodal and versatile video understanding system.
\newblock \emph{arXiv preprint arXiv:2304.14407}.

\bibitem[{Wang et~al.(2023{\natexlab{c}})Wang, Xu, Lan, Hu, Lan, Lee, and Lim}]{wang2023plan}
Lei Wang, Wanyu Xu, Yihuai Lan, Zhiqiang Hu, Yunshi Lan, Roy Ka-Wei Lee, and Ee-Peng Lim. 2023{\natexlab{c}}.
\newblock \href {https://doi.org/10.18653/v1/2023.acl-long.147} {Plan-and-solve prompting: Improving zero-shot chain-of-thought reasoning by large language models}.
\newblock In \emph{Proceedings of the 61st Annual Meeting of the Association for Computational Linguistics (Volume 1: Long Papers)}, pages 2609--2634, Toronto, Canada. Association for Computational Linguistics.

\bibitem[{Wang et~al.(2023{\natexlab{d}})Wang, Zhao, Do, Agarwal, Lee, and Sun}]{wang2023vamos}
Shijie Wang, Qi~Zhao, Minh~Quan Do, Nakul Agarwal, Kwonjoon Lee, and Chen Sun. 2023{\natexlab{d}}.
\newblock Vamos: Versatile action models for video understanding.
\newblock \emph{arXiv preprint arXiv:2311.13627}.

\bibitem[{Wang et~al.(2023{\natexlab{e}})Wang, Sun, Li, Ouyang, Wu, Zhang, Li, and Wang}]{wang2023gpt}
Shuhe Wang, Xiaofei Sun, Xiaoya Li, Rongbin Ouyang, Fei Wu, Tianwei Zhang, Jiwei Li, and Guoyin Wang. 2023{\natexlab{e}}.
\newblock Gpt-ner: Named entity recognition via large language models.
\newblock \emph{arXiv preprint arXiv:2304.10428}.

\bibitem[{Wang et~al.(2023{\natexlab{f}})Wang, Wei, Schuurmans, Le, Chi, Narang, Chowdhery, and Zhou}]{wang2022self}
Xuezhi Wang, Jason Wei, Dale Schuurmans, Quoc Le, Ed~Chi, Sharan Narang, Aakanksha Chowdhery, and Denny Zhou. 2023{\natexlab{f}}.
\newblock Self-consistency improves chain of thought reasoning in language models.
\newblock In \emph{ICLR}.

\bibitem[{Wang et~al.(2021)Wang, Bertasius, Oh, Gupta, Hoai, and Torresani}]{wang2021supervoxel}
Yang Wang, Gedas Bertasius, Tae-Hyun Oh, Abhinav Gupta, Minh Hoai, and Lorenzo Torresani. 2021.
\newblock Supervoxel attention graphs for long-range video modeling.
\newblock In \emph{Proceedings of the IEEE/CVF Winter Conference on Applications of Computer Vision}, pages 155--166.

\bibitem[{Wang et~al.(2022{\natexlab{a}})Wang, Li, Li, He, Huang, Zhao, Zhang, Xu, Liu, Wang et~al.}]{wang2022internvideo}
Yi~Wang, Kunchang Li, Yizhuo Li, Yinan He, Bingkun Huang, Zhiyu Zhao, Hongjie Zhang, Jilan Xu, Yi~Liu, Zun Wang, et~al. 2022{\natexlab{a}}.
\newblock Internvideo: General video foundation models via generative and discriminative learning.
\newblock \emph{arXiv preprint arXiv:2212.03191}.

\bibitem[{Wang et~al.(2022{\natexlab{b}})Wang, Li, Xu, Zhou, Lei, Lin, Wang, Yang, Zhu, Hoiem et~al.}]{wang2022language}
Zhenhailong Wang, Manling Li, Ruochen Xu, Luowei Zhou, Jie Lei, Xudong Lin, Shuohang Wang, Ziyi Yang, Chenguang Zhu, Derek Hoiem, et~al. 2022{\natexlab{b}}.
\newblock Language models with image descriptors are strong few-shot video-language learners.
\newblock \emph{Advances in Neural Information Processing Systems}, 35:8483--8497.

\bibitem[{Wang et~al.(2023{\natexlab{g}})Wang, Sung, Cheng, Bertasius, and Bansal}]{wang2023unified}
Ziyang Wang, Yi-Lin Sung, Feng Cheng, Gedas Bertasius, and Mohit Bansal. 2023{\natexlab{g}}.
\newblock Unified coarse-to-fine alignment for video-text retrieval.
\newblock \emph{arXiv preprint arXiv:2309.10091}.

\bibitem[{Wei et~al.(2022)Wei, Wang, Schuurmans, Bosma, Xia, Chi, Le, Zhou et~al.}]{wei2022chain}
Jason Wei, Xuezhi Wang, Dale Schuurmans, Maarten Bosma, Fei Xia, Ed~Chi, Quoc~V Le, Denny Zhou, et~al. 2022.
\newblock Chain-of-thought prompting elicits reasoning in large language models.
\newblock \emph{Advances in Neural Information Processing Systems}, 35:24824--24837.

\bibitem[{Wu et~al.(2019)Wu, Feichtenhofer, Fan, He, Krahenbuhl, and Girshick}]{wu2019long}
Chao-Yuan Wu, Christoph Feichtenhofer, Haoqi Fan, Kaiming He, Philipp Krahenbuhl, and Ross Girshick. 2019.
\newblock Long-term feature banks for detailed video understanding.
\newblock In \emph{Proceedings of the IEEE/CVF Conference on Computer Vision and Pattern Recognition}, pages 284--293.

\bibitem[{Wu et~al.(2022)Wu, Li, Mangalam, Fan, Xiong, Malik, and Feichtenhofer}]{wu2022memvit}
Chao-Yuan Wu, Yanghao Li, Karttikeya Mangalam, Haoqi Fan, Bo~Xiong, Jitendra Malik, and Christoph Feichtenhofer. 2022.
\newblock Memvit: Memory-augmented multiscale vision transformer for efficient long-term video recognition.
\newblock In \emph{Proceedings of the IEEE/CVF Conference on Computer Vision and Pattern Recognition}, pages 13587--13597.

\bibitem[{Xiao et~al.(2021)Xiao, Shang, Yao, and Chua}]{xiao2021next}
Junbin Xiao, Xindi Shang, Angela Yao, and Tat-Seng Chua. 2021.
\newblock Next-qa: Next phase of question-answering to explaining temporal actions.
\newblock In \emph{Proceedings of the IEEE/CVF conference on computer vision and pattern recognition}, pages 9777--9786.

\bibitem[{Xiao et~al.(2023)Xiao, Yao, Li, and Chua}]{xiao2023can}
Junbin Xiao, Angela Yao, Yicong Li, and Tat~Seng Chua. 2023.
\newblock Can i trust your answer? visually grounded video question answering.
\newblock \emph{arXiv preprint arXiv:2309.01327}.

\bibitem[{Xiao et~al.(2022{\natexlab{a}})Xiao, Yao, Liu, Li, Ji, and Chua}]{xiao2021video}
Junbin Xiao, Angela Yao, Zhiyuan Liu, Yicong Li, Wei Ji, and Tat-Seng Chua. 2022{\natexlab{a}}.
\newblock Video as conditional graph hierarchy for multi-granular question answering.
\newblock In \emph{Proceedings of the 36th AAAI Conference on Artificial Intelligence (AAAI)}, pages 2804--2812.

\bibitem[{Xiao et~al.(2022{\natexlab{b}})Xiao, Zhou, Chua, and Yan}]{xiao2022video}
Junbin Xiao, Pan Zhou, Tat-Seng Chua, and Shuicheng Yan. 2022{\natexlab{b}}.
\newblock Video graph transformer for video question answering.
\newblock In \emph{European Conference on Computer Vision}, pages 39--58. Springer.

\bibitem[{Yang et~al.(2021)Yang, Miech, Sivic, Laptev, and Schmid}]{yang2021just}
Antoine Yang, Antoine Miech, Josef Sivic, Ivan Laptev, and Cordelia Schmid. 2021.
\newblock Just ask: Learning to answer questions from millions of narrated videos.
\newblock In \emph{Proceedings of the IEEE/CVF international conference on computer vision}, pages 1686--1697.

\bibitem[{Yang et~al.(2022{\natexlab{a}})Yang, Miech, Sivic, Laptev, and Schmid}]{yang2022frozenbilm}
Antoine Yang, Antoine Miech, Josef Sivic, Ivan Laptev, and Cordelia Schmid. 2022{\natexlab{a}}.
\newblock Zero-shot video question answering via frozen bidirectional language models.
\newblock In \emph{NeurIPS}.

\bibitem[{Yang et~al.(2022{\natexlab{b}})Yang, Miech, Sivic, Laptev, and Schmid}]{yang2022zero}
Antoine Yang, Antoine Miech, Josef Sivic, Ivan Laptev, and Cordelia Schmid. 2022{\natexlab{b}}.
\newblock Zero-shot video question answering via frozen bidirectional language models.
\newblock \emph{Advances in Neural Information Processing Systems}, 35:124--141.

\bibitem[{Yang et~al.(2020)Yang, Zhu, Wang, Yi, Zadeh, and Morency}]{yang2020gives}
Jianing Yang, Yuying Zhu, Yongxin Wang, Ruitao Yi, Amir Zadeh, and Louis-Philippe Morency. 2020.
\newblock What gives the answer away? question answering bias analysis on video qa datasets.
\newblock \emph{arXiv preprint arXiv:2007.03626}.

\bibitem[{Yang et~al.(2023)Yang, Chu, Feiszli, Goyal, Torresani, and Tran}]{yang2023relational}
Xitong Yang, Fu-Jen Chu, Matt Feiszli, Raghav Goyal, Lorenzo Torresani, and Du~Tran. 2023.
\newblock Relational space-time query in long-form videos.
\newblock In \emph{Proceedings of the IEEE/CVF Conference on Computer Vision and Pattern Recognition}, pages 6398--6408.

\bibitem[{Yao et~al.(2022)Yao, Zhao, Yu, Du, Shafran, Narasimhan, and Cao}]{yao2022react}
Shunyu Yao, Jeffrey Zhao, Dian Yu, Nan Du, Izhak Shafran, Karthik Narasimhan, and Yuan Cao. 2022.
\newblock React: Synergizing reasoning and acting in language models.
\newblock \emph{arXiv preprint arXiv:2210.03629}.

\bibitem[{Ye et~al.(2023)Ye, Xu, Xu, Ye, Yan, Zhou, Wang, Hu, Shi, Shi et~al.}]{ye2023mplug}
Qinghao Ye, Haiyang Xu, Guohai Xu, Jiabo Ye, Ming Yan, Yiyang Zhou, Junyang Wang, Anwen Hu, Pengcheng Shi, Yaya Shi, et~al. 2023.
\newblock mplug-owl: Modularization empowers large language models with multimodality.
\newblock \emph{arXiv preprint arXiv:2304.14178}.

\bibitem[{Yu et~al.(2023)Yu, Cho, Yadav, and Bansal}]{yu2023self}
Shoubin Yu, Jaemin Cho, Prateek Yadav, and Mohit Bansal. 2023.
\newblock Self-chained image-language model for video localization and question answering.
\newblock \emph{NeurIPS}.

\bibitem[{Yu et~al.(2019)Yu, Xu, Yu, Yu, Zhao, Zhuang, and Tao}]{yu2019activitynet}
Zhou Yu, Dejing Xu, Jun Yu, Ting Yu, Zhou Zhao, Yueting Zhuang, and Dacheng Tao. 2019.
\newblock Activitynet-qa: A dataset for understanding complex web videos via question answering.
\newblock In \emph{Proceedings of the AAAI Conference on Artificial Intelligence}, volume~33, pages 9127--9134.

\bibitem[{Zeng et~al.(2022)Zeng, Attarian, Ichter, Choromanski, Wong, Welker, Tombari, Purohit, Ryoo, Sindhwani, Lee, Vanhoucke, and Florence}]{zeng2022socraticmodels}
Andy Zeng, Maria Attarian, Brian Ichter, Krzysztof Choromanski, Adrian Wong, Stefan Welker, Federico Tombari, Aveek Purohit, Michael Ryoo, Vikas Sindhwani, Johnny Lee, Vincent Vanhoucke, and Pete Florence. 2022.
\newblock Socratic models: Composing zero-shot multimodal reasoning with language.
\newblock \emph{arXiv}.

\bibitem[{Zhang et~al.(2021)Zhang, Gupta, and Zisserman}]{zhang2021temporal}
Chuhan Zhang, Ankush Gupta, and Andrew Zisserman. 2021.
\newblock Temporal query networks for fine-grained video understanding.
\newblock In \emph{Proceedings of the IEEE/CVF Conference on Computer Vision and Pattern Recognition}, pages 4486--4496.

\bibitem[{Zhang et~al.(2022)Zhang, Zhang, Li, and Smola}]{zhang2022automatic}
Zhuosheng Zhang, Aston Zhang, Mu~Li, and Alex Smola. 2022.
\newblock Automatic chain of thought prompting in large language models.
\newblock \emph{arXiv preprint arXiv:2210.03493}.

\bibitem[{Zhao et~al.(2023)Zhao, Misra, Kr{\"a}henb{\"u}hl, and Girdhar}]{zhao2023learning}
Yue Zhao, Ishan Misra, Philipp Kr{\"a}henb{\"u}hl, and Rohit Girdhar. 2023.
\newblock Learning video representations from large language models.
\newblock In \emph{Proceedings of the IEEE/CVF Conference on Computer Vision and Pattern Recognition}, pages 6586--6597.

\bibitem[{Zhou et~al.(2023)Zhou, Sch{\"a}rli, Hou, Wei, Scales, Wang, Schuurmans, Cui, Bousquet, Le et~al.}]{zhou2022least}
Denny Zhou, Nathanael Sch{\"a}rli, Le~Hou, Jason Wei, Nathan Scales, Xuezhi Wang, Dale Schuurmans, Claire Cui, Olivier Bousquet, Quoc Le, et~al. 2023.
\newblock Least-to-most prompting enables complex reasoning in large language models.

\bibitem[{Zhou et~al.(2022)Zhou, Muresanu, Han, Paster, Pitis, Chan, and Ba}]{zhou2022large}
Yongchao Zhou, Andrei~Ioan Muresanu, Ziwen Han, Keiran Paster, Silviu Pitis, Harris Chan, and Jimmy Ba. 2022.
\newblock Large language models are human-level prompt engineers.
\newblock \emph{arXiv preprint arXiv:2211.01910}.

\end{thebibliography}
\bibliographystyle{acl_natbib}

\clearpage
\newpage
\appendix
\clearpage
\setcounter{page}{1}

Our appendix consists of Additional Datasets and Metrics (Section~\ref{sec:addtional_datasets}), Qualitative Analysis (Section~\ref{sec:qualitative_analysis}), Additional Implementation Details (Section~\ref{sec:addtional_implementation_details}) and Additional Analysis (Section~\ref{sec:additional_analysis}).

\section{Additional Datasets and Metrics}
\label{sec:addtional_datasets}
In this section, we provide detailed information about the datasets and the metrics we use.
\begin{itemize}
    \item \textbf{NExT-QA~\cite{xiao2021next}} contains 5,440 videos with an average duration of 44s and 48K multi-choice questions and 52K open-ended questions. There are 3 different question types: Temporal, Causal, and Descriptive. Following common practice, we perform zero-shot evaluation on the validation set, which contains 570 videos and 5K multiple-choice questions.
    \item \textbf{IntentQA~\cite{li2023intentqa}} contains 4,303 videos and 16K multiple-choice question-answer pairs focused on reasoning about people's intent in the video. We perform a zero-shot evaluation on the test set containing 2K questions.
    \item \textbf{NExT-GQA~\cite{xiao2023can}} is an extension of NExT-QA with 10.5K temporal grounding annotations associated with the original QA pairs. The dataset was introduced to study whether the existing LVQA models can temporally localize video segments needed to answer a given question. We evaluate all methods on the test split, which contains 990 videos with 5,553 questions, each accompanied by a temporal grounding label. The metrics we used include:  1) Intersection over Prediction (IoP)~\cite{xiao2023can}, which measures whether the predicted temporal window lies inside the ground truth temporal segment, 2) temporal Intersection over Union (IoU), and 3) Acc@GQA, which depicts the percentage of accurately answered and grounded predictions. For IoP and IoU, we report the mean values and values with the overlap thresholds of 0.5.
\end{itemize}

\section{Qualitative Analysis}
\label{sec:qualitative_analysis}
\subsection{Visual Captioners} 
\begin{figure*}[h]
    \centering
    \begin{subfigure}{2\columnwidth}
      \centering
      \includegraphics[width=\linewidth]{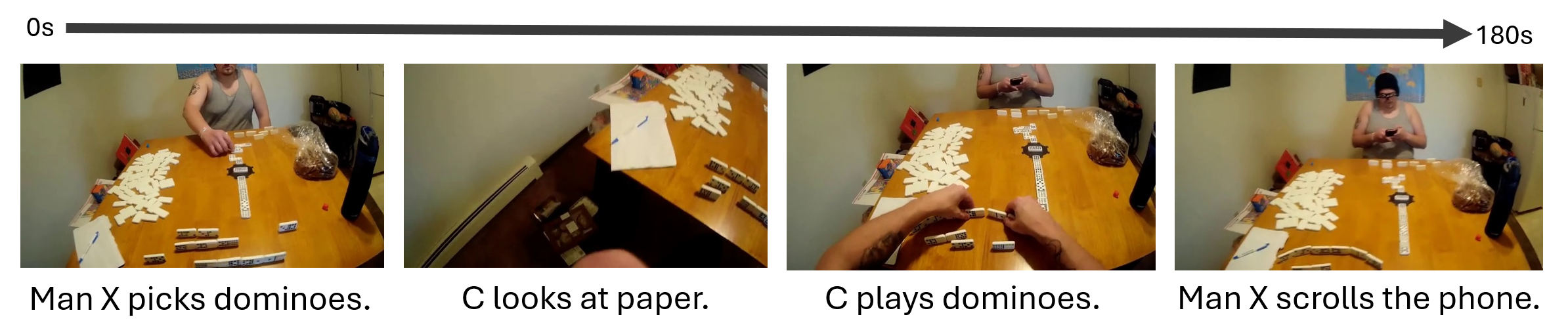}
      \caption{In this 3-minute video, the camera wearer interacts with a man. The camera wearer is always labelled with 'C' and the man is always labelled as 'X'. Man X appears in the first frame. Even though the video loses track of Man X in the second frame, LaViLa still correctly labels him as 'Man X' in the last frame.}
    \end{subfigure} \vfill
    \begin{subfigure}{2\columnwidth}
    \centering
      \includegraphics[width=\linewidth]{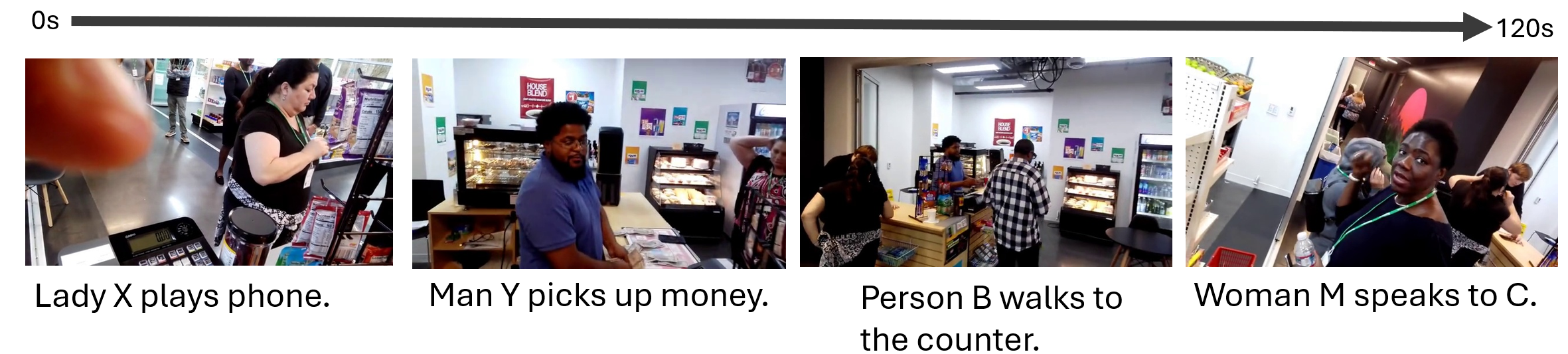}
      \caption{The 2-minute video shows multiple people in a shopping mall. LaViLa labels different people with different characters.}
    \end{subfigure} \vfill
    \begin{subfigure}{2\columnwidth}
    \centering
      \includegraphics[width=\linewidth]{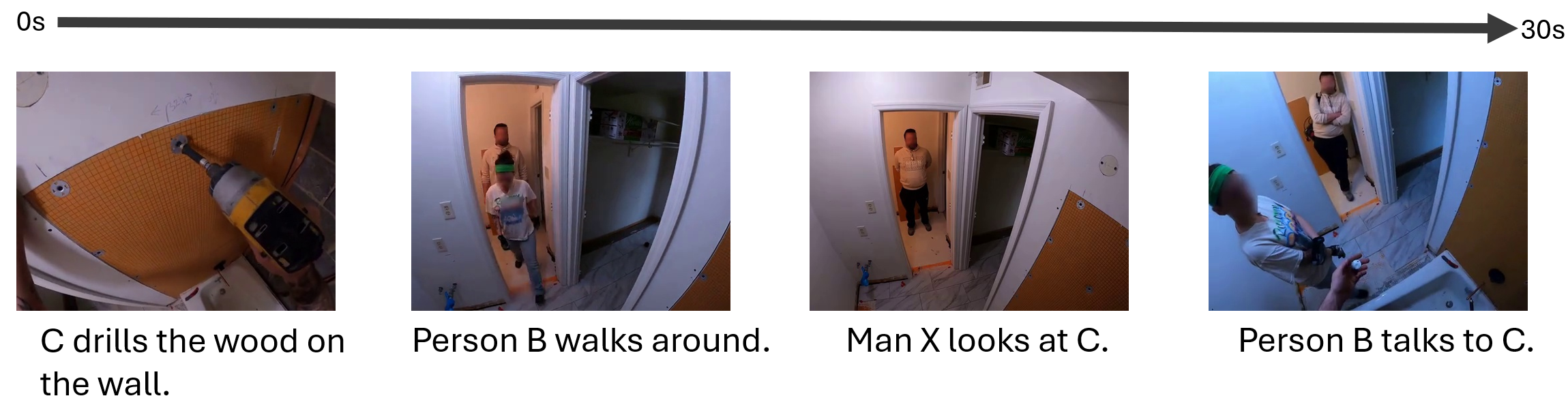}
      \caption{This 30-second video depicts 3 people interacting with each other. Person B appears in the second frame. The thrid frame shows another person interacting with the camera wearer C. Even though person B disappers in the third frame, LaViLa still labels the same entity as Person B in the last frame.}
    \end{subfigure}
\caption{\textbf{Qualitative captioning results on EgoSchema.} 
Our LaViLa visual captioner can differentiate between the camera wearer and other people by assigning the id 'C' to the camera wearer and other ids (e.g., 'B', 'M', 'X', 'Y', etc) to other people. This suggests that our framework using the LaViLa captioner has the basic character ReID ability when the video involves simple interactions between people.}
\label{fig:lavila_vis}
\end{figure*}

In Table~\ref{tab:captioner_vis}, we compare different captions generated by BLIP2 and LaViLa on EgoSchema. LaViLa captions are generally more concise than BLIP2 captions, focusing more on the actions, while BLIP2 focuses more on describing the objects. We also observe that LaViLa is better at differentiating the camera wearer from other people in the video. As shown in the second image in Table~\ref{tab:captioner_vis}, LaViLa captions capture the actions of the other people (not just the camera wearer) in the video. In Figure~\ref{fig:lavila_vis}, we also visualize 3  EgoSchema videos by displaying 4 sparsely-sampled frames. We observe that our framework using the LaViLa captioner can: 1) differentiate between the camera wearer and other people in the video, 2) assign different character ids to different people, and 3) re-identify people if the video consists of simple interaction between the camera wearer and other people.

\begin{table*}[h]
\centering
\begin{tabular}{p{1cm}p{3cm}p{3cm}p{3cm}p{3cm}}
\toprule
&
\begin{minipage}{3.5cm}
    \includegraphics[width=3.3cm]{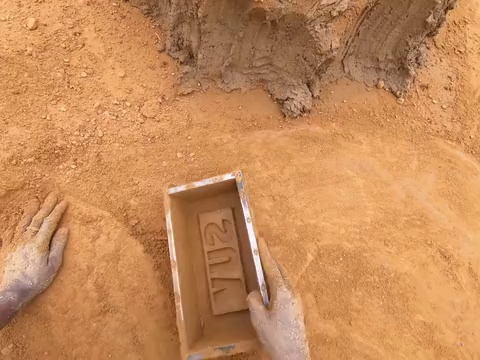}
\end{minipage}
&
\begin{minipage}{3.5cm}
    \includegraphics[width=3.3cm]{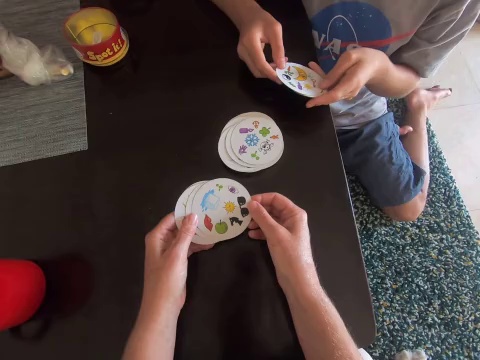}
\end{minipage}
&
\begin{minipage}{3.5cm}
    \includegraphics[width=3.3cm]{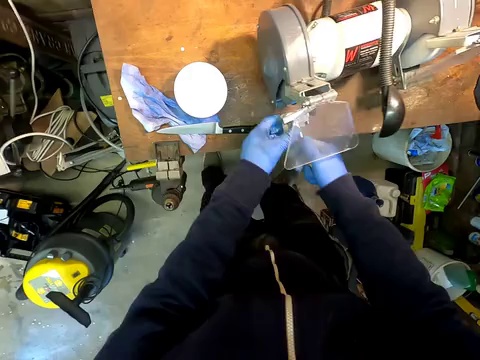}
\end{minipage}
&
\begin{minipage}{3.5cm}
    \includegraphics[width=3.3cm]{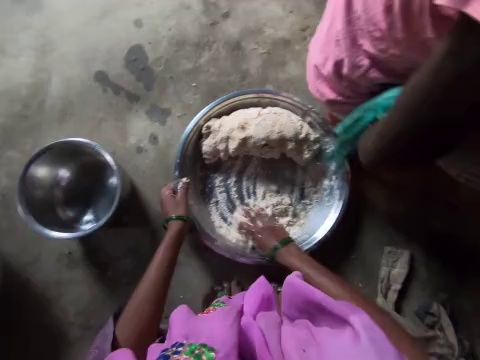}
\end{minipage}
\\
\midrule
\textbf{LaViLa}
&
C drops the brick mould.
&
Man X moves the cards.
&
C puts the cloth on the table.
&
C moves the dough in the tray.
\\
\midrule
\textbf{BLIP2}
&
A person is laying a brick in the dirt.
&
A child is playing a game of monopoly with a tray of paper plates.
&
A person is working on a tool.
&
Woman making dough in a kitchen.
\\
\bottomrule
\end{tabular}
\caption{\textbf{Comparison between different captioners.} \textbf{Top}: frames from EgoSchema videos. \textbf{Middle}: captions generated by LaViLa. \textbf{Bottom}: captions generated by BLIP2. LaViLa captions are more concise than BLIP2 captions. LaViLa is better at differentiating the camera wearer and other people.}
\label{tab:captioner_vis}
\end{table*}

\begin{figure}[h]
    \centering

    \begin{subfigure}{\columnwidth}
      \centering
      \includegraphics[width=1\linewidth]{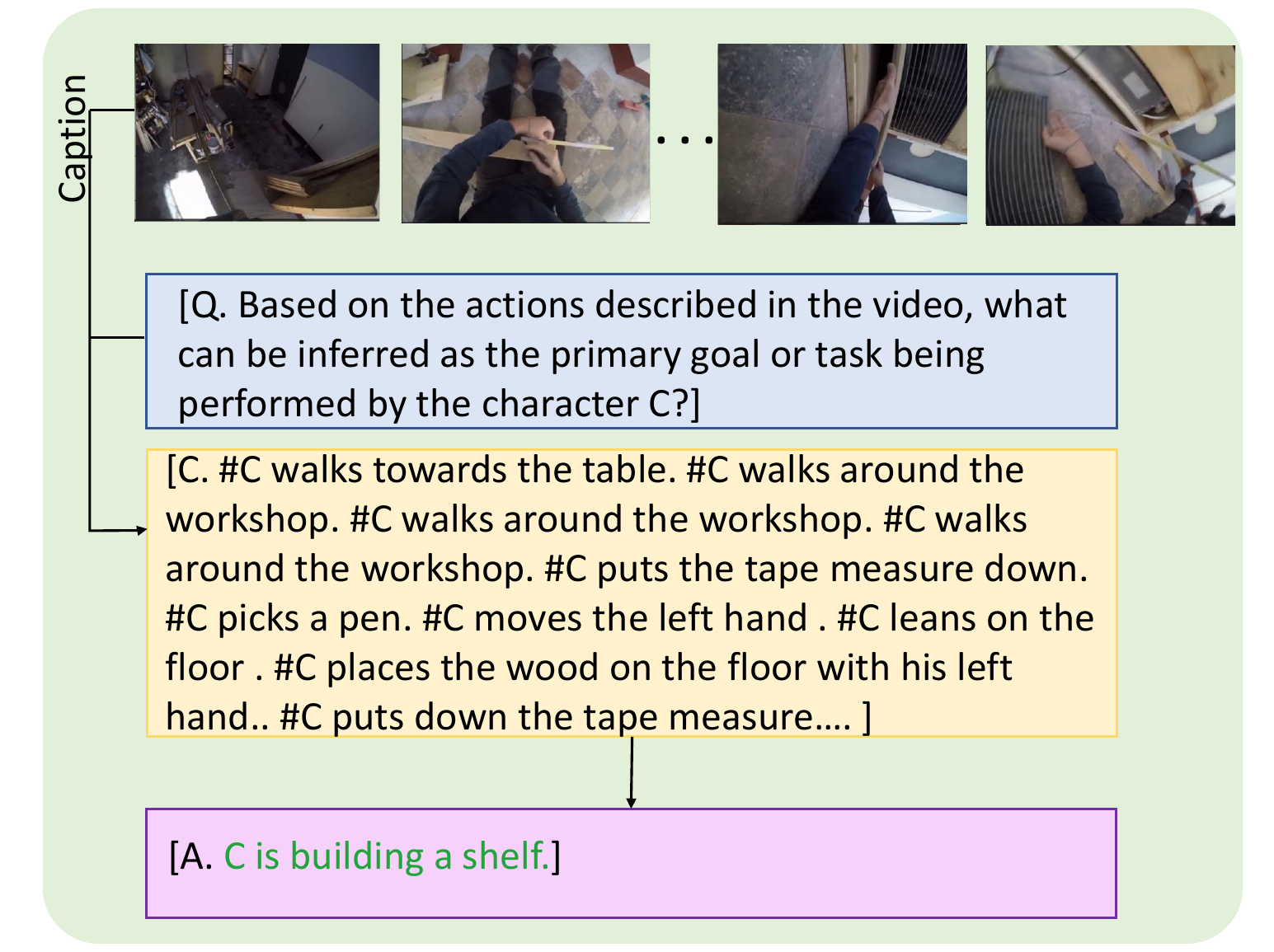}
      \caption{Success case}
      \label{fig:base_a}
    \end{subfigure}
    \vfill%
    \begin{subfigure}{\columnwidth}
      \centering
      \includegraphics[width=\linewidth]{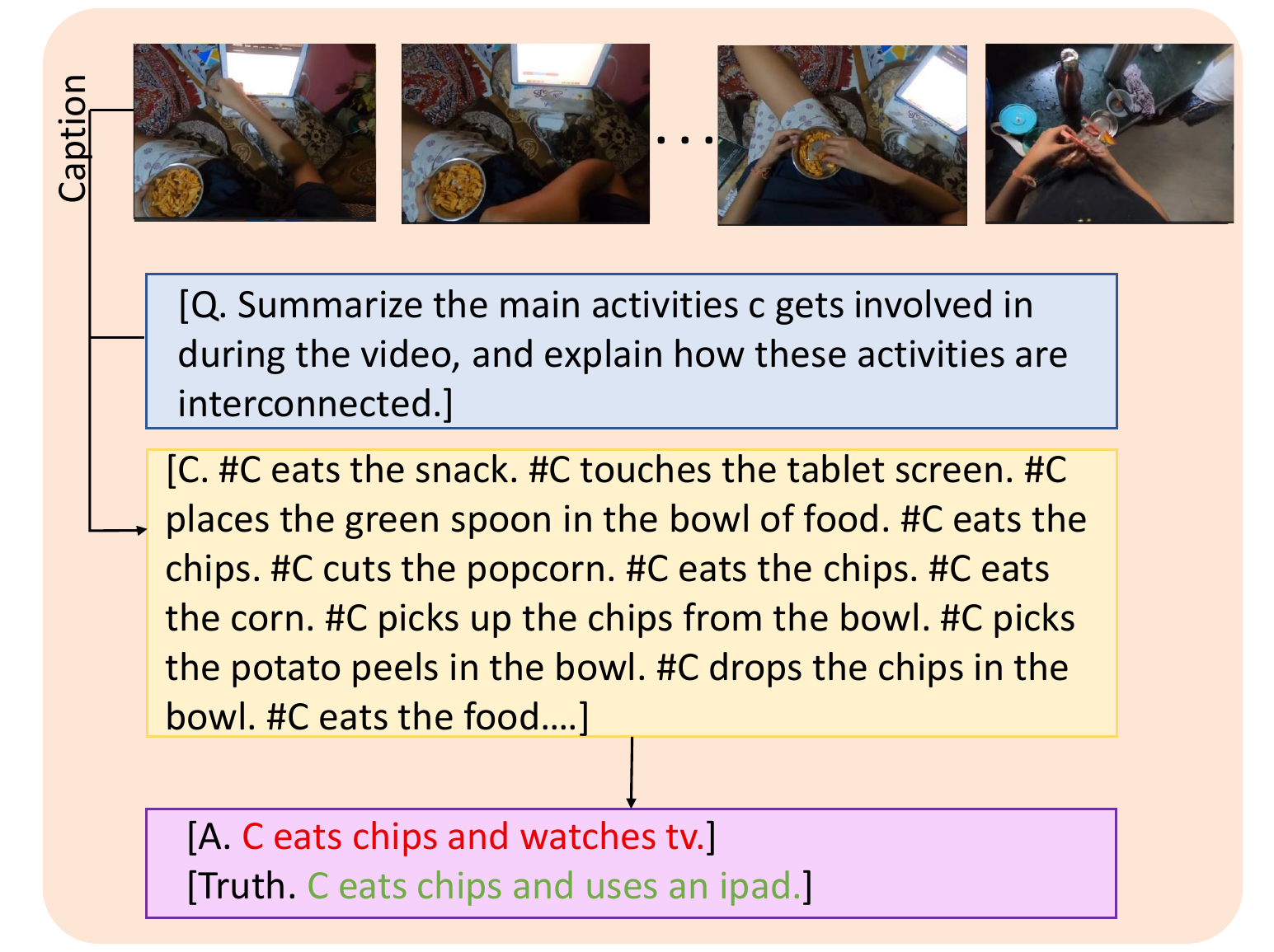}
      \caption{Failure case}
      \label{fig:base_b}
    \end{subfigure}
    \caption{\textbf{Examples of our framework with a standard prompt on EgoSchema.} We show two examples, a successful one (a) and a failed one (b).}
    \label{fig:base}
\end{figure}

\subsection{LLoVi with Standard Prompt}
We show two examples of our method with standard prompt, including a successful one and a failed one in Figure \ref{fig:base}. Our method performs long-range modeling from short-term video captions through LLM to understand the video. In the success case demonstrated in Subfigure~\ref{fig:base_a}, the captions describe the camera wearer's action in a short period of time, such as the interation with the tape measure and the wood. With the short-term captions, LLM understand the long video and answers the question correctly.

\noindent
In the failure case shown in Subfigure~\ref{fig:base_b}, although the video captioner identifies the object in the video correctly as a tablet, LLM understands the action of the camera wearer as watching TV rather than using an iPad. This might be caused by misguidance from the redundant captions that are not related to the question.

\begin{figure}[h]
    \centering

    \begin{subfigure}{\columnwidth}
      \centering
      \includegraphics[width=1\linewidth]{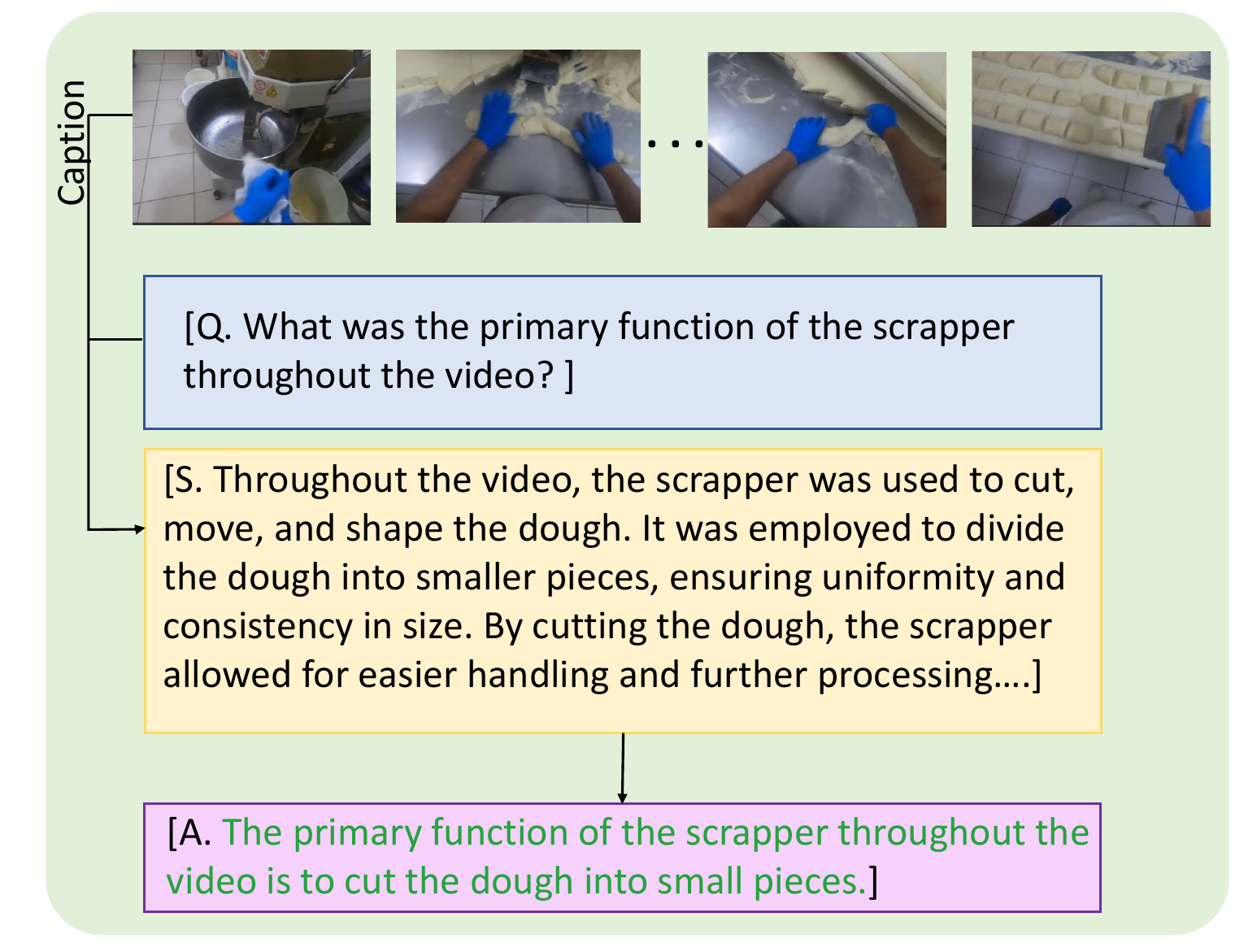}
      \caption{}
      \label{fig:succeed_a}
    \end{subfigure}\hfill%
    \begin{subfigure}{\columnwidth}
    \centering
      \includegraphics[width=\linewidth]{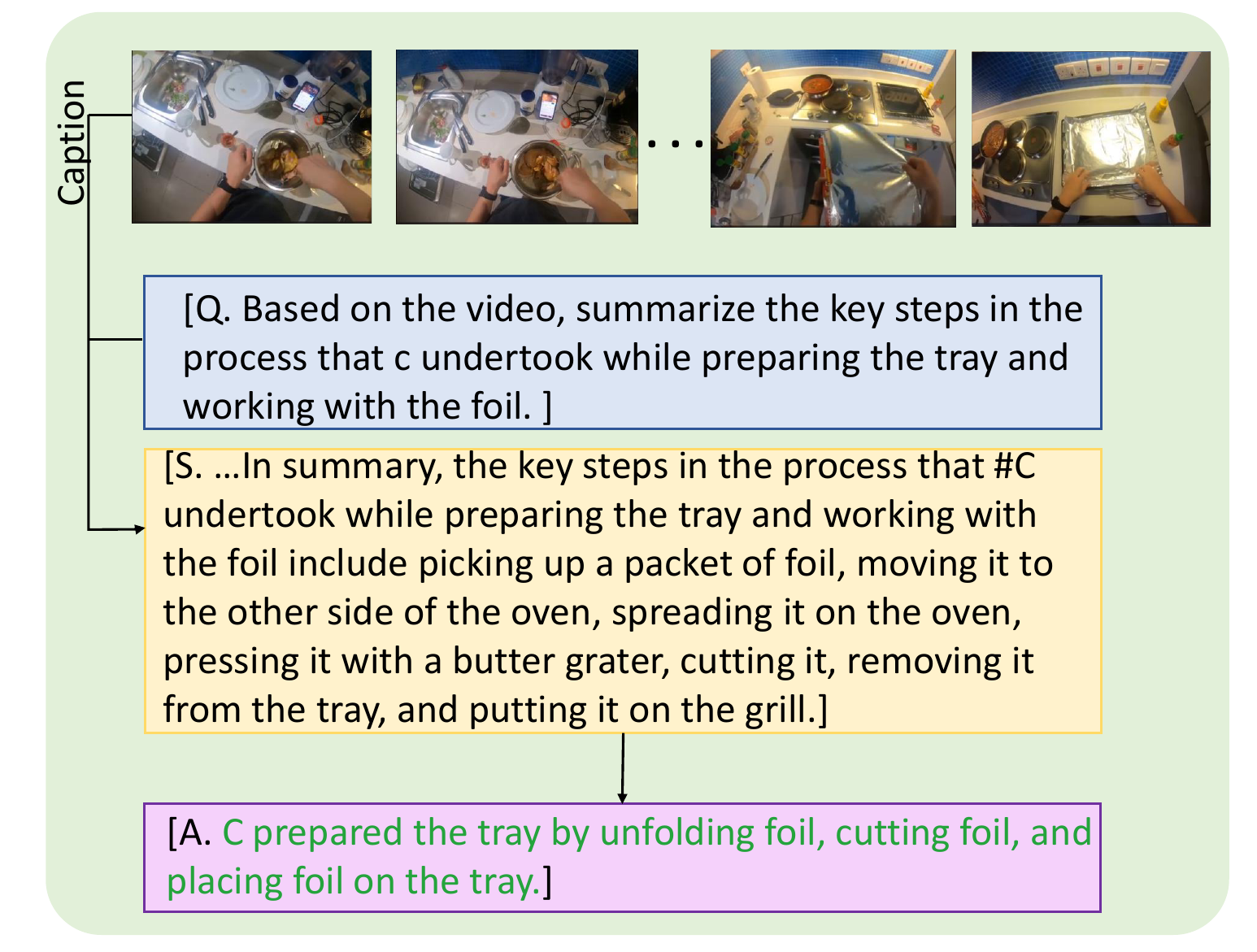}
      \caption{}
      \label{fig:succeed_b}
    \end{subfigure}
    \caption{\textbf{Success cases of our multi-round summarization-based prompt.}}
    \label{fig:succeed}
\end{figure}

\begin{figure}[h]
    \centering

    \begin{subfigure}{\columnwidth}
    \centering
      \includegraphics[width=1\linewidth]{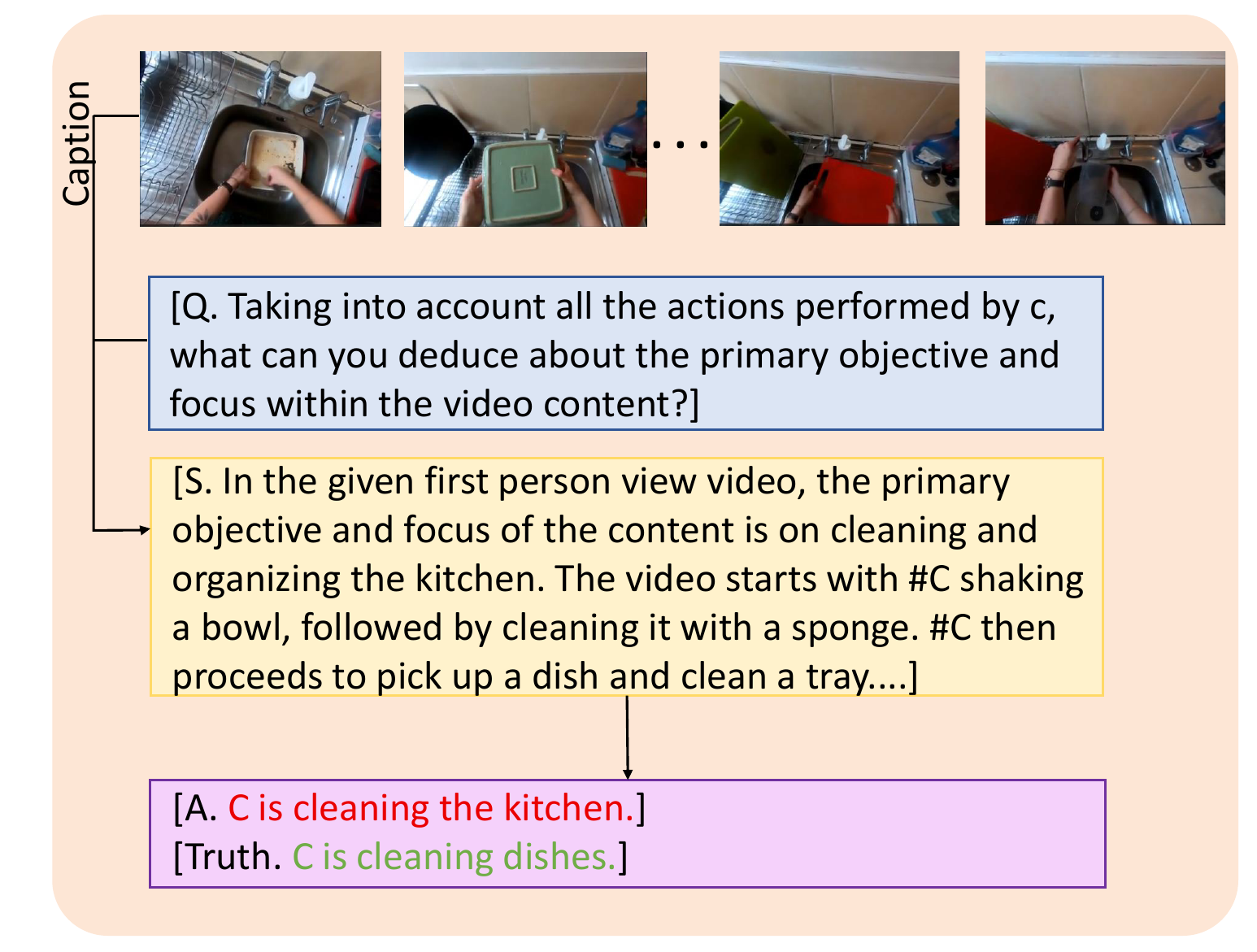}
      \caption{}
      \label{fig:fail_a}
    \end{subfigure}\hfill%
    \begin{subfigure}{\columnwidth}
    \centering
      \includegraphics[width=\linewidth]{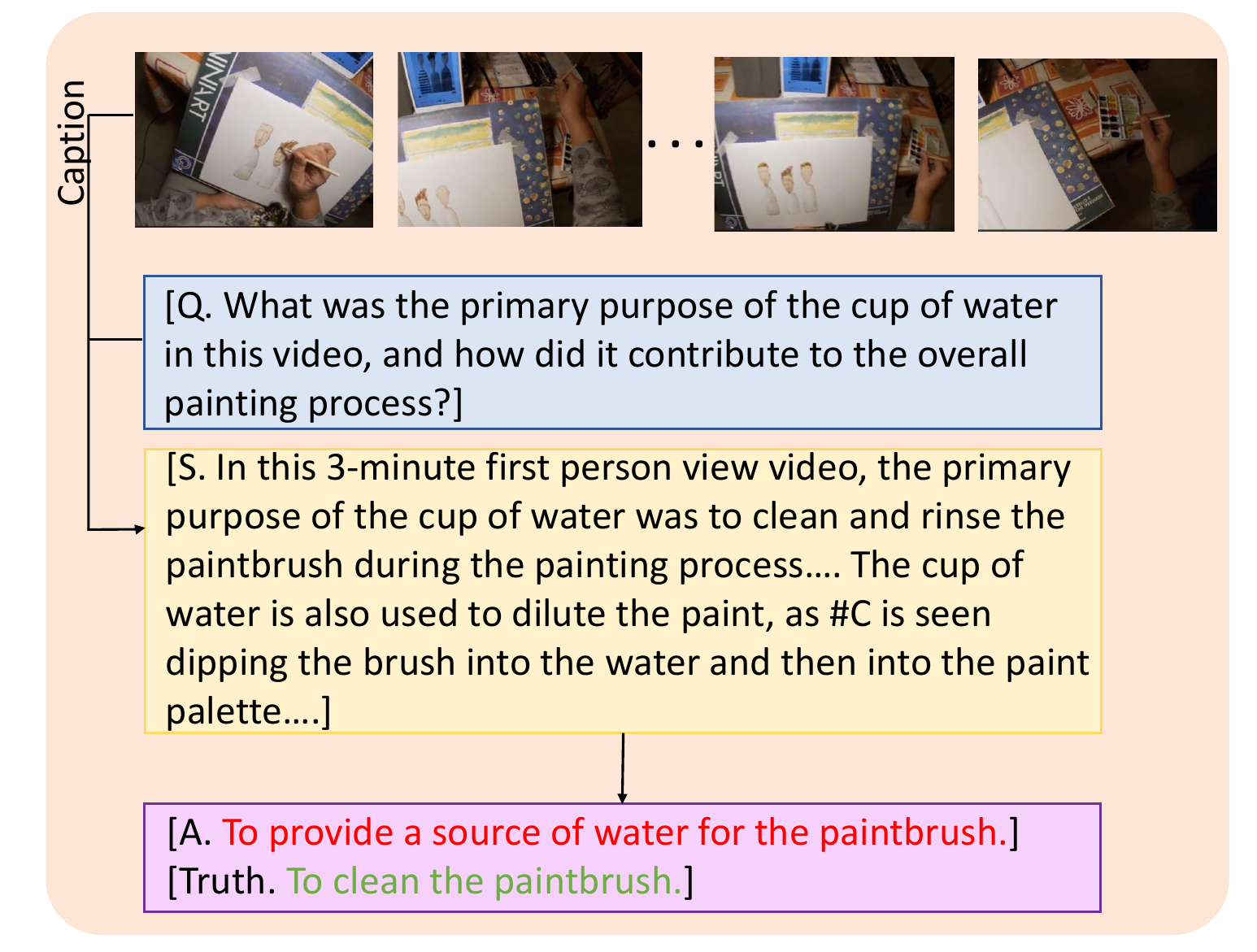}
      \caption{}
      \label{fig:fail_b}
    \end{subfigure}
    \caption{\textbf{Failure cases of our framework with multi-round summarization-based prompt.}}
    \label{fig:fail}
\end{figure}

\begin{figure}[h]
    \centering

    \begin{subfigure}{\columnwidth}
      \centering
      \includegraphics[width=1\linewidth]{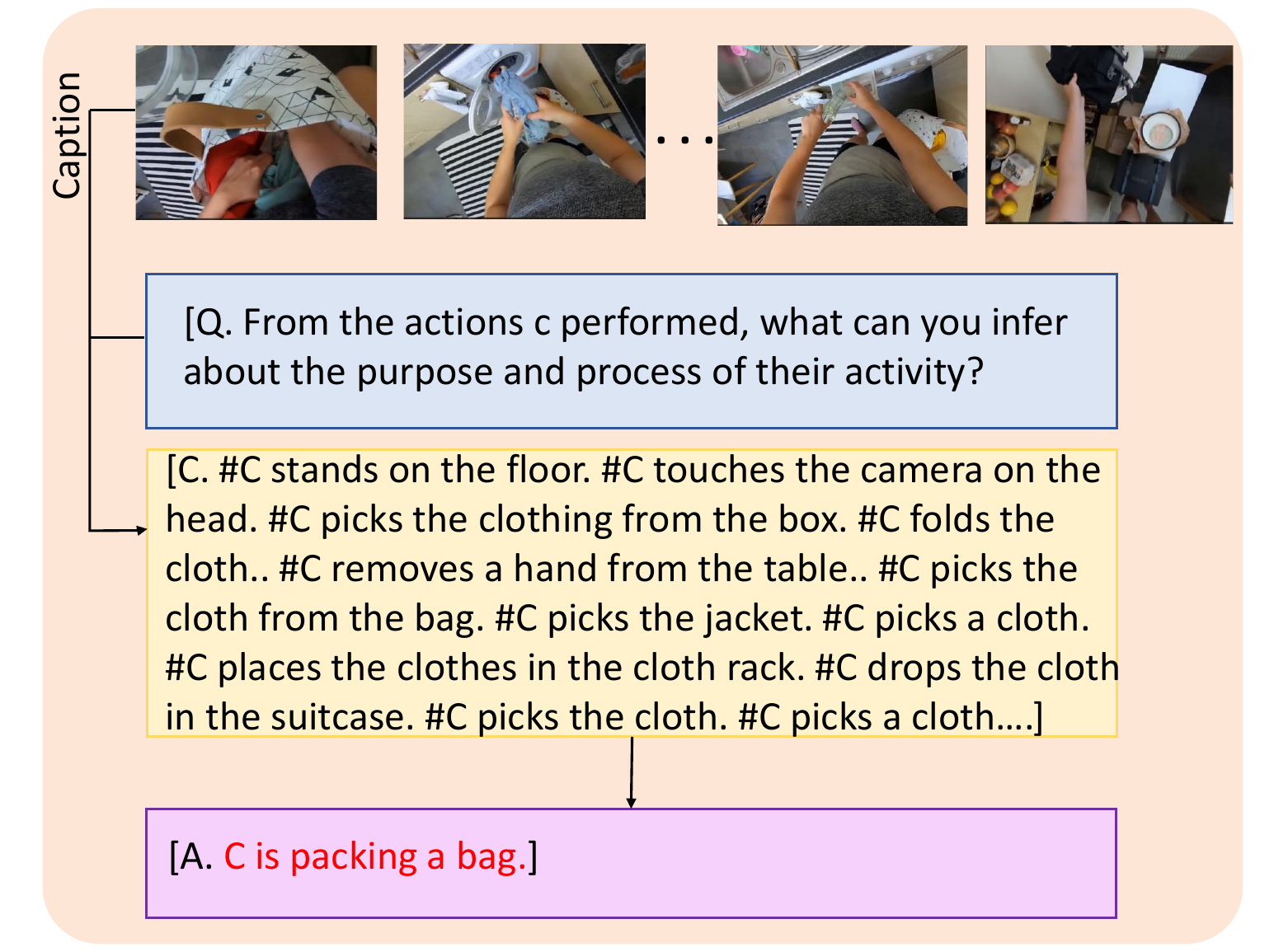}
      \caption{Standard prompt (wrong answer).}
      \label{fig:compare_a}
    \end{subfigure}\hfill%
    \begin{subfigure}{\columnwidth}
      \centering
      \includegraphics[width=\linewidth]{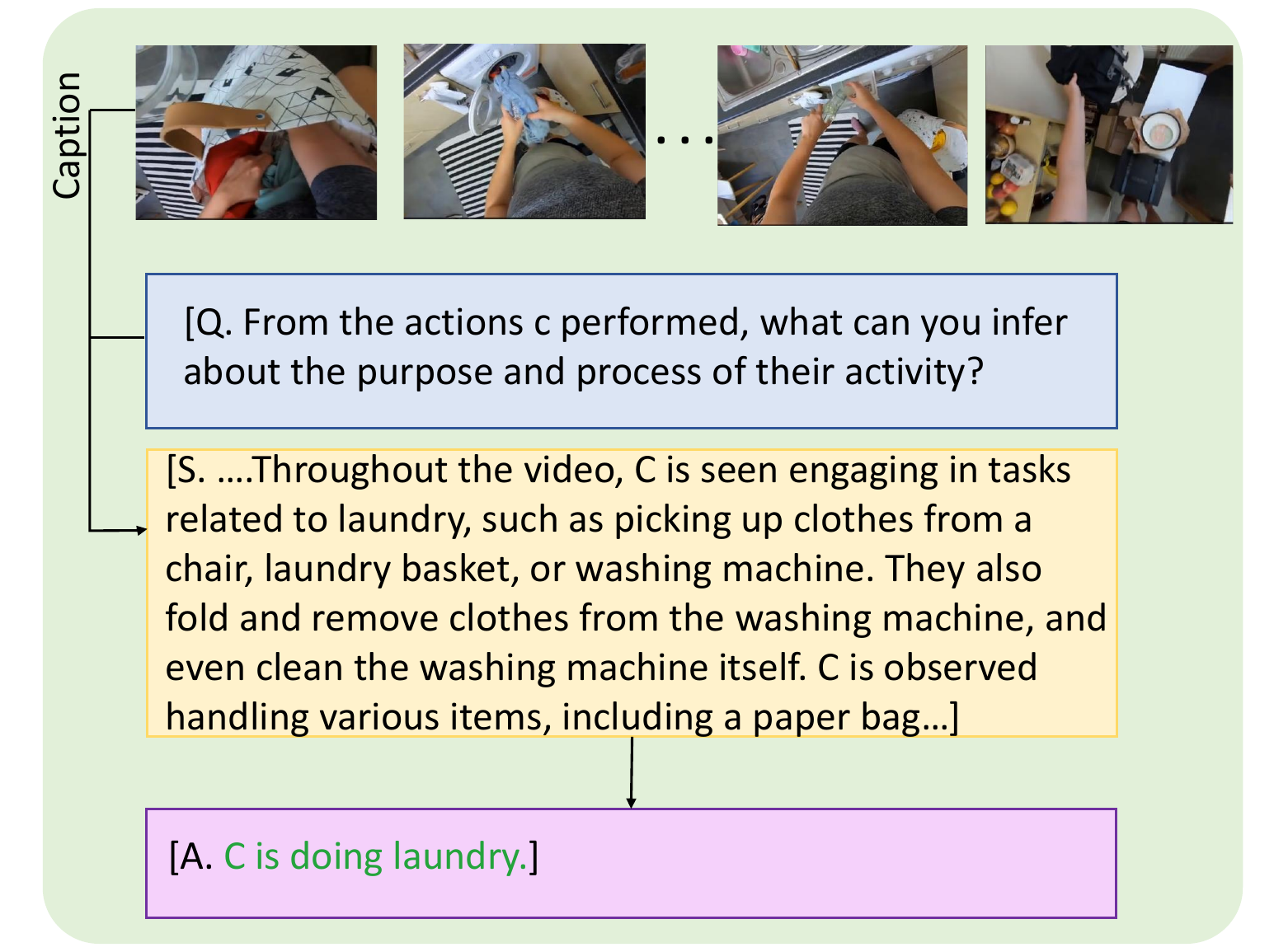}
      \caption{Multi-round summarization-based prompt (correct answer).}
      \label{fig:compare_b}
    \end{subfigure}
    \caption{\textbf{Contrast between our standard prompt and our multi-round summarization-based prompt.} (a) demonstrates the process of answering the question with a standard prompt, and (b) shows answering the question with our multi-round summarization-based prompt.}
    \label{fig:compare}
\end{figure}

\subsection{LLoVi with Multi-round Summarization-based Prompt}
Figure \ref{fig:succeed} illustrates two EgoSchema questions that our framework with multi-round summarization-based prompt answers correctly. In Subfigure~\ref{fig:succeed_a}, the question asks for the primary function of a tool that the video taker uses. However, shown in the first two images, the long video contains descriptions that are not related to the question, such as operating a machine and rolling a dough. As a result, the generated text captions would contain a large section that is not our direction of interest. By summarizing the captions with awareness to the question, LLM extracts key information and cleans redundant captions to provide clearer textual background for answering the question. The same pattern is observed in Subfigure~\ref{fig:succeed_b}.

\noindent
Figure \ref{fig:fail} shows two questions that our method fails to answer. In the summarization stage, the LLM answers the question directly instead of using the question to guide the summarization. For example, in Subfigure~\ref{fig:fail_a}, all the frames show the camera wearer engaging in actions related to washing dishes, but LLM infers that the person is cleaning the kitchen in the summarization stage. This wrong inference further misdirects the following question answering stage, which leads to an incorrect answer. In Subfigure~\ref{fig:fail_b}, LLM concludes that the cup of water is used to dilute the paint because the camera wearer dips the brush into water before dipping it into the paint palette.

\noindent
In Figure \ref{fig:compare}, we also show a question which the standard prompt fails to answer, but the multi-round summarization-based prompt answers correctly. 
In the video in the example question, we observe the camera wearer involving in activities related to laundry, such as picking up clothes from the laundry basket and throwing them into the washing machine. However, the short-term video captions shown in Subfigure~\ref{fig:compare_a} demonstrate the redundancy of actions. The repetitive actions complexes extracting and comprehending the information presented in the caption. For example, excessive captions on picking up clothes can make LLM think that the camera wearer is packing something. Our multi-round summarization-based prompt mitigate this problem by first ask LLM to provide a summary of the captions. The summary shown in Subfigure~\ref{fig:compare_b} states clearly that the camera wearer is doing laundry. With the cleaner and more comprehensive summary, the LLM answer the question correctly.  

\section{Additional Implementation Details}
\label{sec:addtional_implementation_details}
\subsection{Captioners}
For most experiments on EgoSchema, we use LaViLa as the visual captioner. For other pre-trained visual captioners, we use off-the-shelf pre-trained models. Specifically, we use the \texttt{Salesforce/blip2-flan-t5-xl} variant for BLIP-2~\cite{li2023blip2}, \texttt{llava-hf/llava-1.5-13b-hf} variant for LLaVA~\cite{liu2023llava}.

LaViLa is trained on the Ego4D dataset. The original LaViLa train set has 7743 videos with 3.9M video-text pairs and the validation set has 828 videos with 1.3M video-text pairs. 
The EgoSchema dataset is cropped from Ego4D.
Since EgoSchema is designed for zero-shot evaluation and the original LaViLa train set includes EgoSchema videos, we retrain LaViLa on Ego4D videos that do not have any overlap with EgoSchema videos to avoid unfair comparison with other methods. 
After removing the EgoShema videos, the train set consists 6100 videos with 2.3M video-text pairs, and the validation set has 596 videos with 0.7M video-text pairs. We retrain LaViLa on this reduced train set to prevent data leakage. LaViLa training consists of two stages: 1) dual-encoder training and 2) narrator training. Below we provide more details.

\noindent
\textbf{Dual-encoder.}  We use TimeSformer~\cite{bertasius2021space} base model as the visual encoder and a 12-layer Transformer as the text encoder. The input to the visual encoder comprises 4 RGB frames of size $224\times224$. We randomly sample 4 frames from the input video clip and use RandomResizedCrop for data augmentation. The video-language model follows a dual-encoder architecture as CLIP~\cite{radford2021learning} and is trained contrastively. Following LaViLa~\cite{zhao2023learning}, we use 1024 as batch size. We train at a $3\times10^{-5}$ learning rate for 5 epochs on 32 NVIDIA RTX 3090 GPUs.

\noindent
\textbf{Narrator} is a visually conditioned autoregressive Language Model. It consists of a visual encoder, a resampler module, and a text encoder. We use the visual encoder (TimeSformer~\cite{bertasius2021space} base model) from the pretrained dual-encoder (See the previous paragraph). The resampler module takes as input a variable number of video features from the visual encoder and produces a fixed number of visual tokens (i.e. 256). The text decoder is the pretrained GPT-2~\cite{radford2019language} base model with a cross-attention layer inserted in each transformer block which attends to the visual tokens of the resampler module. We freeze the visual encoder and the text decoder, while only training the cross-attention layers of the decoder and the resampler module. Following the design in LaViLa~\cite{zhao2023learning}, we use a batch size of 256 and a learning rate of $3\times10^{-5}$. We use AdamW optimizer~\cite{Kingma2014AdamAM} with $(\beta_1 , \beta_2) = (0.9, 0.999)$ and weight decay 0.01. We train the model on 8 NVIDIA RTX 3090 GPUs for 5 epochs. 

\noindent
\textbf{Narrating video clips.} We use nucleus sampling~\cite{holtzman2019curious} with
$p = 0.95$ and return $K = 5$ candidate outputs. Then we take the narration with the largest confidence score as the final caption of the video clip.

For NExT-QA, we explore CogAgent and LLaVA-1.5 as the visual captioner. For IntentQA and NExT-GQA datasets, we use CogAgent as the visual captioner because of its good performance on NExT-QA. Specifically, we use the \texttt{liuhaotian/llava-v1.5-7b} LLaVA-1.5 variant from Huggingface with the prompt \textit{``USER: $<$image$>$. Describe the image. ASSISTANT: ''}, and the \texttt{THUDM/cogagent-chat-hf} CogAgent variant with the prompt \textit{``$<$image$>$. Describe the image.''}.

\subsection{LLMs}

For most experiments on EgoSchema we use GPT-3.5 as the LLM. Specifically, we use the \texttt{gpt-3.5-turbo-1106} variant. We use 0 as temperature for all experiments.

\noindent
We use \texttt{meta-llama/Meta-Llama-3-8B-Instruct} and \texttt{meta-llama/Meta-Llama-3-70B-Instruct} variants from Huggingface as Llama-3 models. For all Llama3 models, we use greedy sampling to generate the output.

\noindent
For NExT-QA, IntentQA and NExT-GQA datasets, we use GPT-4 as the LLM with the variant \texttt{gpt-4-1106-preview}.

\subsection{Prompting Techniques Implementation}
\label{sec:prompttechniques}
\textbf{Prompt Details.}
We provide detailed prompts for our standard prompt in Table~\ref{tab:llovi_baseline_prompt}, multi-round summarization-based prompt in Table~\ref{tab:llovi_sum_prompt}, Zero-shot Chain of Thought in Table~\ref{tab:llovi_cot}, and Plan-and-Solve prompting in Table~\ref{tab:llovi_plan_solve}. The prompt for the grounded LVQA benchmark is shown in Table~\ref{tab:nextgqa_prompt}.

\begin{table}[h]
\centering
\begin{minipage}{0.99\columnwidth}\vspace{20mm}    
    \centering
    \begin{tcolorbox} 
        \centering
        \hspace{-6mm}
        \begin{tabular}{p{0.99\columnwidth}}
        \hspace{1mm}
        \begin{minipage}{0.99\columnwidth}
        \textbf{User} \\
        Please provide a single-letter answer (A, B, C, D, E) to the following multiple-choice question, and your answer must be one of the letters (A, B, C, D, or E). You must not provide any other response or explanation. \\
        You are given some language descriptions of a first person view video. The video is 3 minute long. Each sentence describes a \blue{\texttt{clip\_length}} clip. Here are the descriptions: \blue{\texttt{Captions}} \\
        You are going to answer a multiple choice question based on the descriptions, and your answer should be a single letter chosen from the choices. \\
        Here is the question: \blue{\texttt{Question}} \\
        Here are the choices. A: \blue{\texttt{Option-A}}. B: \blue{\texttt{Option-B}}. C: \blue{\texttt{Option-C}}. D: \blue{\texttt{Option-D}}. E: \blue{\texttt{Option-E}}. \\
        In your response, the first character should be your answer to this multiple choice question. \\
        \rule[0.25\baselineskip]{\textwidth}{1pt}
        \textbf{Assistant} \\
        \blue{\texttt{Answer}}
        \end{minipage}
        \end{tabular}
    \end{tcolorbox}
    \vspace{-2mm}
    \caption{\textbf{LLoVi with Standard Prompt on EgoSchema.}}
    \label{tab:llovi_baseline_prompt}
\end{minipage}
\end{table}

\begin{table}[h]\centering
\begin{minipage}{0.99\columnwidth}\vspace{0mm}    
    \centering
    \begin{tcolorbox} 
        \centering
        \hspace{-6mm}
        \begin{tabular}{p{0.99\columnwidth}}
        \hspace{1mm}
        \begin{minipage}{0.99\columnwidth}
        \textbf{User} \\
        You are given some language descriptions of a first person view video. Each video is 3 minute long. Each sentence describes a \blue{\texttt{clip\_length}} clip. Here are the descriptions: \blue{\texttt{Captions}} \\
        Please give me a \blue{\texttt{num\_words}} summary. When doing summarization, remember that your summary will be used to answer this multiple choice question: \blue{\texttt{Question}}. \\
        \rule[0.25\baselineskip]{\textwidth}{1pt}
        \textbf{Assistant} \\
        \blue{\texttt{Summary}}
        \end{minipage}
        \end{tabular}
    \end{tcolorbox}
    
    \begin{tcolorbox} 
        \centering
        \hspace{-6mm}
        \begin{tabular}{p{0.99\columnwidth}}
        \hspace{1mm}
        \begin{minipage}{0.99\columnwidth}
        \textbf{User} \\
        Please provide a single-letter answer (A, B, C, D, E) to the following multiple-choice question, and your answer must be one of the letters (A, B, C, D, or E). You must not provide any other response or explanation. \\
        You are given some language descriptions of a first person view video. The video is 3 minute long. Here are the descriptions: \blue{\texttt{Summary}} \\
        You are going to answer a multiple choice question based on the descriptions, and your answer should be a single letter chosen from the choices. \\
        Here is the question: \blue{\texttt{Question}} \\
        Here are the choices. A: \blue{\texttt{Option-A}}. B: \blue{\texttt{Option-B}}. C: \blue{\texttt{Option-C}}. D: \blue{\texttt{Option-D}}. E: \blue{\texttt{Option-E}}. \\
        In your response, the first character should be your answer to this multiple choice question. \\
        \rule[0.25\baselineskip]{\textwidth}{1pt}
        \textbf{Assistant} \\
        \blue{\texttt{Answer}}
        \end{minipage}
        \end{tabular}
    \end{tcolorbox}
    \vspace{-2mm}
    \caption{\textbf{LLoVi with Multi-round Summarization-based Prompt on EgoSchema.} We show the variant (C, Q) $\rightarrow$ S, where we feed the question without potential choices to the summarization stage. \textbf{Top:} caption summarization prompt. \textbf{Bottom:} question answering prompt. In the first stage, GPT3.5 outputs a question-guided summary. In the second stage, GPT3.5 takes the summary without the original captions, then answer the multiple choice question. In practice, we use \textit{num\_words}=500.}
    \label{tab:llovi_sum_prompt}
\end{minipage}
\end{table}

\begin{table}[h]\centering
\begin{minipage}{0.99\columnwidth}\vspace{14mm}    
    \centering
    \begin{tcolorbox} 
        \centering
        \hspace{-6mm}
        \begin{tabular}{p{0.99\columnwidth}}
        \hspace{1mm}
        \begin{minipage}{0.99\columnwidth}
        \textbf{User} \\
        You are given some language descriptions of a first person view video. The video is 3 minute long. Each sentence describes a \blue{\texttt{clip\_length}} clip. Here are the descriptions: \blue{\texttt{Captions}} \\
        You are going to answer a multiple choice question based on the descriptions, and your answer should be a single letter chosen from the choices. \\
        Here is the question: \blue{\texttt{Question}} \\
        Here are the choices. A: \blue{\texttt{Option-A}}. B: \blue{\texttt{Option-B}}. C: \blue{\texttt{Option-C}}. D: \blue{\texttt{Option-D}}. E: \blue{\texttt{Option-E}}. \\
        Before answering the question, let's think step by step. \\
        \rule[0.25\baselineskip]{\textwidth}{1pt}
        \textbf{Assistant} \\
        \blue{\texttt{Answer}} and \blue{\texttt{Rationale}} \\
        \rule[0.25\baselineskip]{\textwidth}{1pt}
        \textbf{User} \\
        Please provide a single-letter answer (A, B, C, D, E) to the multiple-choice question, and your answer must be one of the letters (A, B, C, D, or E). You must not provide any other response or explanation. Your response should only contain one letter. \\
        \rule[0.25\baselineskip]{\textwidth}{1pt}
        \textbf{Assistant} \\
        \blue{\texttt{Answer}}
        \end{minipage}
        \end{tabular}
    \end{tcolorbox}
    \vspace{-2mm}
    \caption{\textbf{LLoVi with Zero-shot Chain of Thought Prompting on EgoSchema.}}
    \label{tab:llovi_cot}
\end{minipage}
\end{table}

\begin{table}[h]\centering
\begin{minipage}{0.99\columnwidth}\vspace{0mm}    
    \centering
    \begin{tcolorbox} 
        \centering
        \hspace{-6mm}
        \begin{tabular}{p{0.99\columnwidth}}
        \hspace{1mm}
        \begin{minipage}{0.99\columnwidth}
        \textbf{User} \\
        You are given some language descriptions of a first person view video. The video is 3 minute long. Each sentence describes a \blue{\texttt{clip\_length}} clip. Here are the descriptions: \blue{\texttt{Captions}} \\
        You are going to answer a multiple choice question based on the descriptions, and your answer should be a single letter chosen from the choices. \\
        Here is the question: \blue{\texttt{Question}} \\
        Here are the choices. A: \blue{\texttt{Option-A}}. B: \blue{\texttt{Option-B}}. C: \blue{\texttt{Option-C}}. D: \blue{\texttt{Option-D}}. E: \blue{\texttt{Option-E}}. \\
        To answer this question, let's first prepare relevant information and decompose it into 3 sub-questions. Then, let's answer the sub-questions one by one. Finally, let's answer the multiple choice question. \\
        \rule[0.25\baselineskip]{\textwidth}{1pt}
        \textbf{Assistant} \\
        \blue{\texttt{Sub-questions}} and \blue{\texttt{Sub-answers}}\\
        \rule[0.25\baselineskip]{\textwidth}{1pt}
        \textbf{User} \\
        Please provide a single-letter answer (A, B, C, D, E) to the multiple-choice question, and your answer must be one of the letters (A, B, C, D, or E). You must not provide any other response or explanation. Your response should only contain one letter. \\
        \rule[0.25\baselineskip]{\textwidth}{1pt}
        \textbf{Assistant} \\
        \blue{\texttt{Answer}}
        \end{minipage}
        \end{tabular}
    \end{tcolorbox}
    \vspace{-2mm}
    \caption{\textbf{LLoVi with Plan-and-Solve Prompting on EgoSchema.}}
    \label{tab:llovi_plan_solve}
\end{minipage}
\vspace{14mm}
\end{table}

\begin{table}[h]\centering
\begin{minipage}{0.99\columnwidth}\vspace{20mm}    
    \centering
    \begin{tcolorbox} 
        \centering
        \hspace{-6mm}
        \begin{tabular}{p{0.99\columnwidth}}
        \hspace{1mm}
        \begin{minipage}{0.99\columnwidth}
        \textbf{User} \\
        I will provide video descriptions and one question about the video. The video is 1 FPS and the descriptions are the captions every 2 frames. Each caption starts with the frame number.To answer this question, what is the minimun frame interval to check? Follow this format: [frame\_start\_index, frame\_end\_index]. Do not provide any explanation. \\
        Here are the descriptions: \blue{\texttt{Captions}} \\
        Here is the question: \blue{\texttt{Question}} \\
        Please follow the output format as follows: \#Example1: [5, 19]. \#Example2: [30, 60]. \#Example3: [1, 10] and [50, 60] \\
        \rule[0.25\baselineskip]{\textwidth}{1pt}
        \textbf{Assistant} \\
        \blue{\texttt{Answer}}
        \end{minipage}
        \end{tabular}
    \end{tcolorbox}
    \vspace{-2mm}
    \caption{\textbf{LLoVi Prompt on NExT-GQA.}}
    \label{tab:nextgqa_prompt}
\end{minipage}
\end{table}

\noindent
\textbf{Output Processing.}
When answering multiple choice questions, GPT3.5 usually outputs complete sentences instead of a single-letter answer, i.e. A, B, C, D, or E. 
One way to obtain the single-character response is to perform post-processing on the output, which usually requires substantial engineering efforts.
In our work, however, we observe that GPT3.5 is very sensitive to the starting sentences of the prompts. Therefore, we explicitly prompt it as in Table~\ref{tab:llovi_baseline_prompt} to force GPT3.5 to generate a single character as response. In practice, we take out the first character of the output as the final answer.

\section{Additional Analysis}
\label{sec:additional_analysis}

In this section, we provide additional analysis on the EgoSchema Subset using the standard prompt.

\subsection{Additional Ablations on NExT-QA}
In Table ~\ref{tab:nextqa_more}, we show our framework's performance using different combinations of the visual captioners and the LLMs. Specifically, we explore BLIP-2, LLaVA-1.5, CogAgent as the visual captioner, and Llama-3-70B, GPT-3.5, GPT-4 as the LLM. We notice that the best results are achieved by the combination of CogAgent and GPT-4. For all LLMs, CogAgent constantly outperforms LLaVA-1.5, and LLaVA-1.5 constantly outperforms BLIP-2. Additionally, we observe that GPT-3.5 and Llama-3-70B achieves similar performance, and that they are both significantly outperformed by GPT-4. These results suggest that stronger visual captioners and LLMs always lead to better results under our framework, and that our framework is able to benefits from future development of the visual captioners and the LLMs.

\begin{table}[h]
\centering
\setlength{\tabcolsep}{7pt}
\resizebox{\columnwidth}{!}{
\begin{tabular}{lccccc}
\toprule
Captioner & LLM & C. & T. & D. & All  \\ 
\midrule
BLIP-2 & \multirow{3}{*}{Llama-3-70B} & 62.8 & 53.6 & 68.5 & 60.7 \\
LLaVA-1.5 & & 63.1 & 56.3 & 70.0 & 62.0 \\
CogAgent & & 67.9 & 58.2 & 75.9 & 66.0 \\
\midrule
BLIP-2 & \multirow{3}{*}{GPT-3.5} & 57.9 & 51.1 & 67.1 & 57.2 \\
LLaVA-1.5 & & 59.0 & 53.7 & 68.8 & 58.7 \\
CogAgent & & 67.1 & 60.1 & 76.5 & 66.3 \\
\midrule
BLIP-2 & \multirow{3}{*}{GPT-4} & 67.1 & 57.6 & 73.8 & 65.1\\
LLaVA & & 69.5 & 61.0 & 75.6 & 67.7 \\
CogAgent & & \textbf{73.7} & \textbf{70.2} & \textbf{81.9} & \textbf{73.8} \\
\bottomrule
\end{tabular}
}
\caption{\textbf{Different Captioners and LLMs on NExT-QA.} We observe that CogAgent constantly outperforms LLaVA-1.5, followed by BLIP-2, for all LLMs. GPT-4 constantly outperforms Llama-3-70B and GPT-3.5 for all captioners.}
\label{tab:nextqa_more}
\vspace{-0.1in}
\end{table}

\subsection{Accuracy on Different Question Types}
To better understand the strengths and limitations of our LVQA framework, we manually categorize questions in the EgoSchema Subset into 5 categories: (1) Purpose/Goal Identification, (2) Tools and Materials Usage, (3) Key Action/Moment Detection, (4) Action Sequence Analysis, (5) Character Interaction, and break down our system's performance according to each of the category as shown in Table~\ref{tab:breakdown}. The details description of each question category is shown in Table~\ref{tab:cats}. Note that some questions belong to more than one category. Based on this analysis, we observe that almost half of the questions relate to purpose/goal identification, which makes intuitive sense as inferring human goals/intent typically requires a very long video analysis. We also observe that a significant portion of the questions relate to tool usage, key action detection, and action sequence analysis. Lastly, the smallest fraction of the questions belong to character interaction analysis.

\begin{table}[h]
\centering
\setlength{\tabcolsep}{2pt}
\resizebox{\columnwidth}{!}{
    \begin{tabular}{lcc}
    \toprule
    Question Category & Category Percentage & Acc. \\ 
    \midrule
    Purpose/Goal Identification & 49.2 & 56.5 \\
    Tools and Materials Usage & 21.8 & 53.2 \\
    Key Action/Moment Detection & 21.6 & 53.7 \\
    Action Sequence Analysis & 18.2 & 51.6 \\
    Character Interaction & 9.4 & 63.8  \\
    \bottomrule
    \end{tabular}
}
\caption{\textbf{Accuracy on different question categories of EgoSchema.} We manually categorize each question in the EgoSchema Subset into 5 categories. Note that each question may belong to one or more categories. Our system performs the best on questions that involve character interaction analysis or human purpose/goal identification. This is encouraging as both of these questions typically require a very long-form video analysis.}
\label{tab:breakdown}
\end{table}

\begin{table*}
\centering
\setlength{\tabcolsep}{1pt}
    \begin{tabular}{p{0.15\textwidth}p{0.02\textwidth}p{0.25\textwidth}p{0.02\textwidth}p{0.50\textwidth}}
    \toprule
    Question \par Category & & Description & & Examples \\ 
    \midrule
    Purpose/Goal Identification & & primary goals, intentions, summary, or overarching themes of the video & & 1. Taking into account all the actions performed by c, what can you deduce about the primary objective and focus within the video content? \newline 2. What is the overarching theme of the video, considering the activities performed by both characters?\\
    \midrule
    Tools and Materials Usage & & how the character engages with specific tools, materials, and equipment & & 1. What was the primary purpose of the cup of water in this video, and how did it contribute to the overall painting process? \newline \ 2. Explain the significance of the peeler and the knife in the video and their respective roles in the preparation process. \\
    \midrule
    Key Action / Moment Detection & & identify crucial steps/actions, the influence/rationale of key action/moment/change on the whole task & & 1. Out of all the actions that took place, identify the most significant one related to food preparation and explain its importance in the context of the video. \newline 2. Identify the critical steps taken by c to organize and prepare the engine oil for use on the lawn mower, and highlight the importance of these actions in the overall video narrative. \\
    \midrule
    Action Sequence Analysis & & compare and contrast different action sequences, relationship between different actions, how characters adjust their approach, efficacy and precision, expertise of the character & & 1. What is the primary sequence of actions performed by c throughout the video, and how do these actions relate to the overall task being performed? \newline 2. Considering the sequence of events, what can be inferred about the importance of precision and accuracy in the character's actions, and how is this demonstrated within the video? \\
    \midrule
    Character Interaction & & how characters interact and collaborate, how their roles differ  & & 1. What was the main purpose of the actions performed by both c and the man throughout the video, and how did their roles differ? \newline 2. Describe the general activity in the room and how the different characters and their actions contribute to this environment. \\
    \bottomrule
    \end{tabular}
\caption{\textbf{Question categories of EgoSchema.} We manually categorize each question in the EgoSchema Subset into 5 categories. Note that each question may belong to one or more categories.}
\label{tab:cats}
\end{table*}

In Table ~\ref{tab:breakdown}, we show our system's performance on each of the above-discussed question categories. Our results indicate that our system performs the best in the Character Interaction category (\textbf{63.8\%}). One possible explanation is that the LaViLa model, which we use as our visual captioner, is explicitly pretrained to differentiate the camera wearer from other people, making it well-suited for understanding various interactions between characters in the video. We also observe that our model performs quite well on the remaining categories ($>$\textbf{50\%}). It is especially encouraging to see strong results (\textbf{56.5\%}) in the Purpose/Goal Identification category since inferring human intentions/goals from the video inherently requires very long-form video analysis.


\end{document}